\documentclass{article} 
\usepackage{iclr2027_conference,times}

\usepackage{amsmath,amsfonts,bm}

\def\eqref#1{equation~\ref{#1}}

\def\1{\bm{1}}

\DeclareMathAlphabet{\mathsfit}{\encodingdefault}{\sfdefault}{m}{sl}
\SetMathAlphabet{\mathsfit}{bold}{\encodingdefault}{\sfdefault}{bx}{n}

\DeclareMathOperator*{\argmin}{arg\,min}

\usepackage{hyperref}
\usepackage{url}
\usepackage{graphicx}
\usepackage{multirow}
\usepackage{booktabs}
\usepackage{placeins}
\usepackage{float}

\title{SoFT: Soft Targets for Generalizable LLM Fine-Tuning}

\author {
  \begin{minipage}{\textwidth}
  \centering
    {\bf Huihao Jing}\textsuperscript{ \hspace{-0.2em}\includegraphics[height=1em]{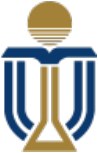}}\footnotemark[1],
    {\bf Wenbin Hu}\textsuperscript{ \hspace{-0.2em}\includegraphics[height=1em]{figures/HKUST.pdf}}\footnotemark[1],
    {\bf Shaojin Chen}\textsuperscript{ \hspace{-0.2em}\includegraphics[height=1em]{figures/HKUST.pdf}},
    {\bf Haochen Shi}\textsuperscript{ \hspace{-0.2em}\includegraphics[height=1em]{figures/HKUST.pdf}},
    {\bf Zhongwei Xie}\textsuperscript{ \hspace{-0.2em}\includegraphics[height=1em]{figures/HKUST.pdf}},
    {\bf Guijia Zhang}\textsuperscript{ \hspace{-0.2em}\includegraphics[height=1em]{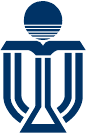}},
    {\bf Yuxuan Liu}\textsuperscript{ \hspace{-0.2em}\includegraphics[height=1em]{figures/HKUST.pdf}},
    {\bf Haoyu Huang}\textsuperscript{ \hspace{-0.2em}\includegraphics[height=1em]{figures/HKUST.pdf}},
    {\bf Haoran Li}\textsuperscript{
      \hspace{-0.2em}
      \includegraphics[height=1em]{figures/HKUST.pdf}
    }\footnotemark[2]\hspace{0.15em},
    {\bf Yangqiu Song}\textsuperscript{ \hspace{-0.2em}\includegraphics[height=1em]{figures/HKUST.pdf}}\\
    \textsuperscript{\includegraphics[height=1em]{figures/HKUST.pdf}}HKUST, 
    \textsuperscript{\includegraphics[height=1em]{figures/HKUST_GZ_mark.pdf}}HKUST-GZ\\
    \texttt{hjingaa@connect.ust.hk}\\
  \end{minipage}
}

\iclrfinalcopy 
\begin{document}
\renewcommand{\thefootnote}{\fnsymbol{footnote}}
\footnotetext[1]{Equal contribution.}
\footnotetext[2]{Corresponding author.}
\renewcommand{\thefootnote}{\arabic{footnote}}
\setcounter{footnote}{0}

\maketitle

\begin{abstract}
Distillation enables student language models to acquire new capabilities from expert teachers. However, integrating knowledge from multi-teacher, multi-domain demonstrations into a single student remains challenging. We study supervised fine-tuning (SFT) in this setting, where students must acquire diverse capabilities while maintaining generalization beyond the training tasks. Our experiments reveal varying trade-offs between in-distribution learning and out-of-distribution generalization across SFT methods, motivating more explicit control over this balance. To this end, we propose soft-target fine-tuning (SoFT) to balance learning from teacher demonstrations with retaining the Base model's existing capabilities. SoFT sets a minimum target probability for each demonstrated token while making the smallest KL change to the Base distribution. The resulting objective couples learning from demonstrations with adaptively weighted regularization toward the Base model. We further use domain-specific gradient budgets to control this balance and determine a probability threshold for each trajectory. Experiments on mixed-domain reasoning and agentic tasks show that SoFT achieves the best overall performance among the compared methods, with improvements in both in-distribution capability acquisition and out-of-distribution generalization. We provide the our data and code in \href{https://github.com/HKUST-KnowComp/SoFT}{this repository}.
\end{abstract}

\begin{figure}[H]
\vspace{-0.2in}
\centering
\begin{minipage}[c]{0.50
\linewidth}
\centering
\raisebox{-0.03\linewidth}{\includegraphics[width=\linewidth]{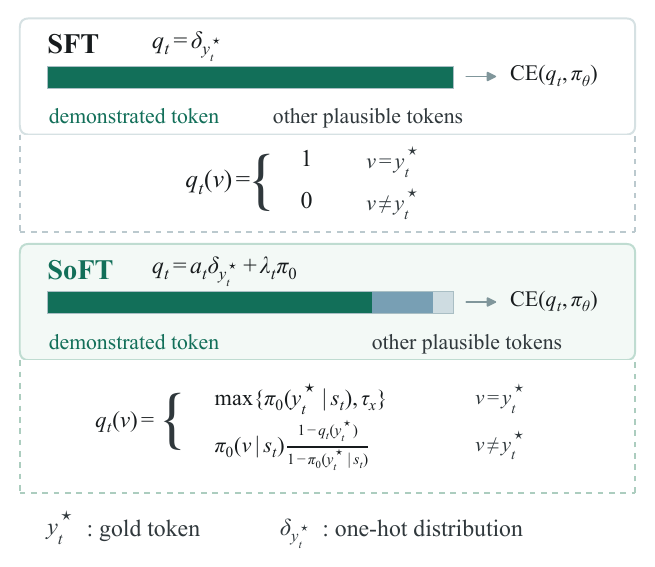}}
\end{minipage}\hfill
\begin{minipage}[c]{0.48\linewidth}
\centering
\includegraphics[width=\linewidth]{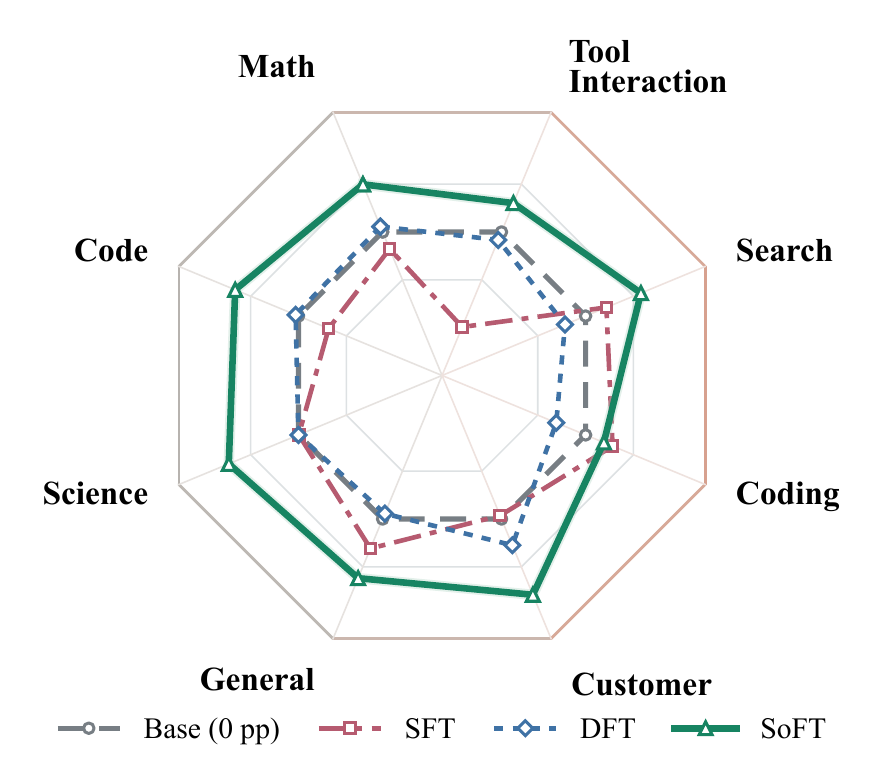}
\end{minipage}
\vspace{-0.17in}
\caption{SoFT overview and mixed-data outcomes. Left: schematic targets; SFT uses a one-hot target, while SoFT retains Base-supported alternatives. Right: changes relative to Base across reasoning and agentic domains.}
\label{fig:domain-overview}
\end{figure}

\section{Introduction}
\label{sec:introduction}

Supervised fine-tuning (SFT) on expert demonstrations is a standard way to transfer new capabilities to a language model~\citep{ouyang2022training,lambert2024tulu3}. Demonstrations may contain intermediate reasoning~\citep{muennighoff2025s1,deepseek2025r1,guha2025openthoughts}, tool decisions, or long multi-turn interactions~\citep{raoof2026data,jung2026procuasft}, and practical corpora often mix trajectories from several teachers and task domains~\citep{wan2024fusellm,guha2025openthoughts}. A single student must then acquire all of these capabilities without losing the useful behavior it learned during pretraining~\citep{li2025diversity,wang2026anchored}.

Standard SFT is poorly suited to this goal. A teacher trajectory is only one of many valid continuations: another teacher, or the student itself, might use different wording, reasoning steps, tool calls, or stopping points~\citep{kim2025ownwords,raoof2026data}. Yet SFT trains toward a one-hot target on every observed token. Repeated updates therefore favor the demonstrated trajectory at the expense of other plausible continuations~\citep{li2025diversity,chen2025limitingconfidence}, especially in long reasoning and agentic trajectories~\citep{twist2026reasoningcollapse,jung2026procuasft}. SFT also ignores the student's starting point, pushing every demonstrated token toward certainty whether Base already predicts it well or finds it unlikely~\citep{chen2025limitingconfidence,wang2026prift}. Under mixed data, these costs are uneven across domains: in Figure~\ref{fig:domain-overview} (right), SFT and DFT improve some domains but fall below Base on others. Existing variants reweight or select training signals~\citep{xia2024less,lin2024rho,chen2025limitingconfidence,wu2026dft,wang2026prift} or anchor the student to its initialization~\citep{li2018l2sp,zhu2025asft,wang2026anchored}, but they typically tune learning and preservation separately (Section~\ref{sec:sft-improvement}).

We propose soft-target fine-tuning (\textbf{SoFT}), which replaces the one-hot target with a soft target (Figure~\ref{fig:domain-overview}, left): the distribution closest to Base in KL that gives the demonstrated token at least a floor probability. This design has three advantages. First, it changes Base minimally: the target adds only the probability mass the floor requires and keeps Base's relative preferences among other tokens, so plausible continuations are not needlessly suppressed. Second, it couples learning and preservation: the target splits into complementary demonstration and Base-preservation weights, so supervision concentrates on tokens that Base poorly supports. Third, it is simple to control. A single gradient budget, defined as the fraction of SFT's initial gradient on demonstrated tokens that SoFT retains, sets the learning strength, and a per-trajectory floor adapts it to the student's initial probabilities. Training needs only offline demonstrations, with no teacher logits or online rollouts. Our contributions are threefold:
\begin{enumerate}
    \item \textbf{Mixed-data SFT hides an acquisition--retention trade-off.} When one student learns from multiple teachers and domains, gains on some task families often come with sharp losses on others. At 7B, four of six fine-tuning baselines fall below Base in aggregate score, and at 8B, SFT gains in-distribution but loses out-of-distribution.
    \item \textbf{SoFT turns the training target itself into the control for this trade-off.} We derive a closed-form, minimum-KL soft target that decomposes exactly into complementary learning and preservation weights, and calibrate it with a per-domain gradient budget. SoFT needs no teacher logits or online rollouts.
    \item \textbf{SoFT gains more while drifting less.} SoFT achieves the best aggregate score in all three settings and improves 22 of 24 benchmark scores over Base, e.g., raising $\tau^2$ OOD from 20.67 to 48.33. Its policy drift is only about one tenth of SFT's, and our analysis shows that stronger imitation of demonstrated tokens alone does not explain these gains.
\end{enumerate}

\section{Preliminary} %
\label{sec:preliminary}

\subsection{Supervised Fine-Tuning}
Let $\mathcal D=\{(x,y^\star)\}$ denote a corpus of expert demonstrations, where $y^\star$ is the complete reference response to query $x$. SFT minimizes the negative log-likelihood of the demonstrated response:
\begin{equation}
\mathcal L_{\mathrm{SFT}}(\theta)
=
\mathbb{E}_{(x,y^\star)\sim\mathcal D}\!\left[
-\log \pi_\theta(y^\star\mid x)
\right].
\label{eq:sft-ce-objective}
\end{equation}

\subsection{Applications of SFT} 
SFT remains a central stage in recent model training pipelines and transfers both reasoning and agentic capabilities from verified demonstrations. For reasoning, supervised rationales expose intermediate computations and decomposition strategies in addition to final answers~\citep{ho2023reasoning,magister2023teaching,hsieh2023distilling,shridhar2023distilling,yue2024mammoth}. Recent recipes such as s1, OpenThoughts, and skill-aware distillation improve this transfer through data curation, scaling, and student-specific selection~\citep{muennighoff2025s1,guha2025openthoughts,zhang2026skillaware}. DeepSeek-R1 and Qwen3 further demonstrate reasoning transfer from stronger models to smaller students~\citep{deepseek2025r1,yang2025qwen3}. For agentic capabilities, trajectories supervise tool use, environment interaction, recovery, and task completion. Kimi K2.5, Kimi K3, Qwen3-Coder-Next, DeepSeek V4.1, OpenThoughts-Agent, and ProCUA-SFT apply this approach to long-horizon and executable tasks~\citep{kimi2026k25,kimi2026k3,cao2026qwen3codernext,deepseek2026v41,raoof2026data,jung2026procuasft}. Since trajectory SFT uses a fixed corpus and requires only teacher outputs, it supports black-box distillation and reuses each verified trajectory across many updates~\citep{kim2016sequence,guha2025openthoughts}. These properties make SFT a scalable first stage for acquiring new model capabilities.

\subsection{SFT Improvement}
\label{sec:sft-improvement}
Many SFT improvements on a fixed demonstration corpus can be written in the general form
\begin{equation}
\mathcal L(\theta)
=
\mathbb{E}_{(x,y^\star)\sim\mathcal D}\!\left[
\sum_t
\left(
-a_t\log\pi_\theta(y_t^\star\mid s_t)
+\lambda_t\Omega_t\!\left(
\pi_\theta(\cdot\mid s_t),
\pi_0(\cdot\mid s_t)
\right)
\right)
\right],
\label{eq:baseline-decomposition}
\end{equation}
where $s_t=(x,y_{<t}^\star)$ is the teacher-forced state, $a_t$ controls how strongly the demonstrated token is learned, and $\lambda_t\Omega_t$ controls the deviation of the updated policy $\pi_\theta$ from the Base policy $\pi_0$. Standard SFT is recovered by setting $a_t=1$ and $\lambda_t=0$.

The first line of work adjusts $a_t$ to reallocate learning strength across examples or tokens. LESS, Rho-1, and PriFT-mass select training signals, while Confidence and DFT assign model-dependent weights~\citep{xia2024less,lin2024rho,wang2026prift,wu2026dft}. These methods change how strongly each demonstration token is learned, but usually retain the one-hot target on $y_t^\star$. The second line adjusts $\lambda_t$ or $\Omega_t$ to limit how far the student moves from the Base model. ASFT adds a KL anchor to DFT~\citep{zhu2025asft}, while Anchored Learning controls distributional drift through an intermediate target~\citep{wang2026anchored}. L2-SP is the parameter-space analogue, replacing the policy discrepancy above with an anchor between $\theta$ and its initialization $\theta_0$~\citep{li2018l2sp}. These methods preserve prior behavior through a regularizer whose strength is usually specified separately from the SFT signal.

SoFT couples these two lines. Its adaptive target is equivalent to jointly setting the demonstration coefficient $a_t$ and the Base-preservation coefficient $\lambda_t$, so that a larger value of one leaves a smaller value for the other. This coupling controls acquisition and preservation through one calibrated target rather than two independently tuned objectives. Sections~\ref{sec:soft-target} and~\ref{sec:rstar} define this target and its calibration.

\section{Method} %
\label{sec:method}

We introduce \emph{soft-target fine-tuning} (SoFT), an offline objective that couples learning from demonstrations with preserving the Base policy. From the demonstrated sequence and the Base distribution alone, SoFT constructs one adaptive training target for each token, without online rollouts or teacher logits.

\subsection{From a Global Budget to Token Targets}
\label{sec:control-overview}

\begin{figure}[H]
\centering
\includegraphics[width=\linewidth]{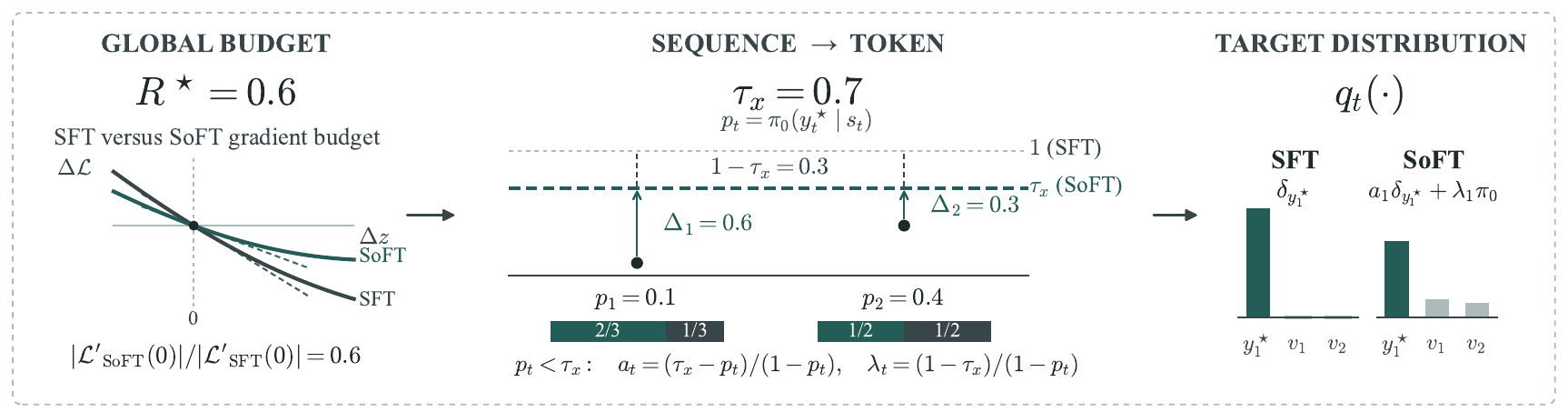}
\vspace{-0.2in}
\caption{From a shared gradient budget (left), to a sequence-specific floor (middle), to token-level SoFT targets (right). Values are illustrative.}
\label{fig:soft-control-flow}
\end{figure}

Figure~\ref{fig:soft-control-flow} gives an overview, from left to right. A global gradient budget $R^\star$ (left) sets SoFT's initial gradient on demonstrated tokens relative to SFT. For each sequence, this budget is converted into a probability floor $\tau_x$ (middle). At each teacher-forced state $s_t=(x,y_{<t}^\star)$, each position whose Base probability $\pi_0(y_t^\star\mid s_t)$ falls below $\tau_x$ is lifted to the floor, which determines a demonstration weight $a_t$; for example, with $\tau_x=0.7$, a token with Base probability $0.1$ receives $a_t=2/3$. The weight then defines a token-level soft target (right) that, unlike the one-hot SFT target, keeps part of Base's probability on alternative tokens. Thus, $R^\star$ is selected globally, $\tau_x$ adapts to each sequence, and the target adapts to each token. Sections~\ref{sec:soft-target}--\ref{sec:rstar} formalize these three steps in reverse order: the target, its decomposition into weights, and the calibration of $\tau_x$.

\subsection{Soft Target Construction} %
\label{sec:soft-target}

At each teacher-forced state $s_t$, SoFT replaces the one-hot target with the distribution $q_t$ closest to the Base policy $\pi_0$ that assigns at least probability $\tau_x$ to the demonstrated token $y_t^\star$ (we leave its dependence on $\tau_x$ implicit):
\begin{equation}
q_t
=
\argmin_{q\in\Delta(\mathcal V)}
\mathrm{KL}\!\left(
q\,\|\,\pi_0(\cdot\mid s_t)
\right)
\quad
\text{s.t.}
\quad
q(y_t^\star)\geq\tau_x.
\label{eq:soft-projection}
\end{equation}
The KL term discourages unnecessary changes to the Base distribution, and the constraint enforces the probability floor. Solving the Lagrangian gives the closed form
\begin{equation}
q_t(y_t^\star) = \max\!\left\{ \pi_0(y_t^\star\mid s_t),\tau_x \right\}, 
\qquad q_t(v) = \pi_0(v\mid s_t) \frac{1-q_t(y_t^\star)} {1-\pi_0(y_t^\star\mid s_t)},
\quad v\neq y_t^\star.
\label{eq:floor-target}
\end{equation}
The solution adds only the probability mass required by the floor and rescales all other tokens proportionally, so Base's relative preferences among them are unchanged. If $\pi_0(y_t^\star\mid s_t)\geq\tau_x$, the constraint is inactive and $q_t=\pi_0(\cdot\mid s_t)$; as $\tau_x\rightarrow1$, the target approaches the one-hot SFT target. Appendix~\ref{app:soft-projection} gives the full derivation.

\subsection{Soft Target Decomposition}
\label{sec:coupled-objective}

We now decompose the soft target to connect it with the learning and preservation terms in Eq.~\ref{eq:baseline-decomposition}. Letting $\delta_{y_t^\star}$ denote the one-hot distribution on $y_t^\star$, Eq.~\ref{eq:floor-target} can be written as
\begin{equation}
a_t
=
\frac{\max\{\tau_x-\pi_0(y_t^\star\mid s_t),\,0\}}{1-\pi_0(y_t^\star\mid s_t)},
\qquad
\lambda_t=1-a_t,
\qquad
q_t
=
a_t\delta_{y_t^\star}
+
\lambda_t\pi_0(\cdot\mid s_t).
\label{eq:soft-coupling}
\end{equation}
Here, $a_t$ weights learning from the demonstrated token and $\lambda_t$ weights preservation of the Base distribution, and the two are complementary by construction. Tokens that Base already predicts above the floor receive $a_t=0$ and keep their Base target; all other tokens receive exactly the extra probability needed to reach $\tau_x$.

Substituting Eq.~\ref{eq:soft-coupling} into the cross-entropy gives
\begin{equation}
\mathcal L_{\mathrm{SoFT}}(\theta)
=
\mathbb{E}_{(x,y^\star)\sim\mathcal D}\!\left[
\sum_t
\left(
-a_t\log\pi_\theta(y_t^\star\mid s_t)
+
\lambda_t
\mathrm{CE}\!\left(
\pi_0(\cdot\mid s_t),
\pi_\theta(\cdot\mid s_t)
\right)
\right)
\right].
\label{eq:soft-loss}
\end{equation}
Since this cross-entropy equals $\mathrm{KL}(\pi_0\|\pi_\theta)$ up to a $\theta$-independent constant, SoFT instantiates Eq.~\ref{eq:baseline-decomposition} with $\Omega_t$ as the forward KL, through a single soft target whose learning and preservation weights are coupled rather than tuned independently.

\subsection{Adaptive Control with \texorpdfstring{$R^\star$}{R*}}
\label{sec:rstar}

The floor $\tau_x$ determines both weights in Eq.~\ref{eq:soft-coupling}. Since the Base probabilities $\pi_0(y_t^\star\mid s_t)$ vary across demonstrations, a floor shared by all sequences would produce different effective learning strengths. We therefore solve for a sequence-specific floor that meets a shared gradient budget $R^\star$.

At the Base initialization, the gradient magnitude on the gold logit (the logit of $y_t^\star$) is $1-\pi_0(y_t^\star\mid s_t)$ for SFT and $a_t(1-\pi_0(y_t^\star\mid s_t))$ for SoFT. We choose $\tau_x$ such that
\begin{equation}
R_x(\tau_x)
=
\frac{\sum_t a_t(1-\pi_0(y_t^\star\mid s_t))}
{\sum_t(1-\pi_0(y_t^\star\mid s_t))}
=
R^\star.
\label{eq:rstar}
\end{equation}
Each numerator term equals $q_t(y_t^\star)-\pi_0(y_t^\star\mid s_t)$, so $R^\star$ is the fraction of Base's gap to the one-hot target that the soft target closes (e.g., $R^\star=0.3$ closes 30\% of it); equivalently, it is the fraction of SFT's initial gold-logit gradient that SoFT retains. As $R^\star$ moves from zero to one, SoFT moves from preserving the Base distribution toward one-hot SFT. Since $R_x(\tau)$ is continuous and monotone in $\tau$, $\tau_x$ is obtained by scalar bisection (Appendix~\ref{app:soft-rstar}). Because $q_t$ depends only on $\pi_0$, it stays fixed during training and can be precomputed; Appendix~\ref{sec:soft-implementation} gives a memory-efficient top-$K$ approximation.

\section{Experimental Setup}
\label{sec:experimental-setup}

We evaluate whether SoFT can learn from mixed training data while preserving generalization beyond the training tasks. We compare SoFT with Base, standard SFT~\citep{ouyang2022training}, Confidence~\citep{chen2025limitingconfidence}, Rho-1~\citep{lin2024rho}, DFT~\citep{wu2026dft}, PriFT-mass~\citep{wang2026prift}, and L2-SP~\citep{li2018l2sp} on reasoning and agentic tasks. Within each setting, all fine-tuned methods share the initialization, data, and training budget, and hyperparameters are selected on held-out validation data.

\subsection{Domain-Aware \texorpdfstring{$R^\star$}{R*}}
\label{sec:domain-rstar}

For SoFT, we select one gradient budget $R_d^\star$ per domain using held-out validation data. Each sequence $x$ then uses the budget of its domain $d(x)$:
\begin{equation}
R_x(\tau_x)=R_{d(x)}^\star,
\label{eq:domain-rstar}
\end{equation}
so that $R_d^\star$ sets the domain-level balance and $\tau_x$ adapts it to each sequence. For General, Math, Science, and Code, Qwen2.5-1.5B-Instruct uses $R_d^\star=(0.15,0.30,0.45,0.60)$ and Qwen2.5-7B uses $(0.25,0.35,0.45,0.55)$. Qwen3-8B uses $(0.3,0.8,0.3,0.3)$ for Customer, Coding, Search, and Tool Interaction.

\subsection{Tasks and Evaluation}

For reasoning, we fine-tune Qwen2.5-1.5B-Instruct and Qwen2.5-7B~\citep{yang2024qwen25} on 36,000 teacher trajectories, equally divided among Math, Code, Science, and General, with 800 validation examples. For agentic tasks, we fine-tune Qwen3-8B~\citep{yang2025qwen3} on 8,252 trajectories across Customer, Coding, Search, and Tool Interaction, with 824 validation examples. Teachers include DeepSeek-R1, GLM-5.1, Kimi-K2.5, QwQ-32B/STILL-2, Qwen3-235B-A22B, GPT-OSS-20B, DeepSeek V4 Flash, and Qwen3.6-27B (Appendix~\ref{app:training-dataset}). Within each capability domain, we use in-distribution (ID) for training-related evaluation and out-of-distribution (OOD) for cross-benchmark evaluation (Table~\ref{tab:evaluation-suites}).

Reasoning evaluations use OpenR1-Math~\citep{lozhkov2025openr1math}, APPS~\citep{hendrycks2021apps}, RiddleSense~\citep{lin2021riddlesense}, GSM8K~\citep{cobbe2021gsm8k}, ARC-Challenge~\citep{clark2018arc}, BoolQ~\citep{clark2019boolq}, and MBPP with EvalPlus tests~\citep{austin2021mbpp,liu2023evalplus}. Science MCQ is our filtered subset of Mixture-of-Thoughts~\citep{openr12025mixture}, not a separate official benchmark. Agentic evaluations use $\tau^2$~\citep{barres2025tau2}, LiveCodeBench~\citep{jain2024livecodebench}, TMax~\citep{ivison2026tmax}, BigCodeBench~\citep{zhuo2024bigcodebench}, BFCL V4~\citep{patil2025bfcl}, and FRAMES~\citep{krishna2024frames}. For $\tau^2$, training and the 60 ID tasks use the retail and airline domains, while the 300 OOD tasks come from telecom and banking knowledge. Search ID uses held-out Nemotron search prompts~\citep{nvidia2025agenticdata}.

\begin{table}[!t]
\centering
\vspace{-0.1in}
\caption{Evaluation suites and case counts before repeated attempts. BigCodeBench counts evaluation cases.}
\label{tab:evaluation-suites}
\vspace{0.07in}
\small
\setlength{\tabcolsep}{3pt}
\renewcommand{\arraystretch}{1.04}
\begin{tabular*}{\linewidth}{@{\extracolsep{\fill}}llll@{}}
\toprule
Setting & Domain & ID (cases) & OOD (cases) \\
\midrule
\multirow{4}{*}{Reasoning} & Math & OpenR1-Math (300) & GSM8K (1,319) \\
 & Code & APPS (250) & MBPP+ (378) \\
 & Science & Science MCQ (1,000) & ARC-Challenge (608) \\
 & General & RiddleSense (1,018) & BoolQ (708) \\
\midrule
\multirow{4}{*}{Agentic} & Customer & $\tau^2$ (60) & $\tau^2$ (300) \\
 & Coding & LiveCodeBench (694) & BigCodeBench (2,280) \\
 & Search & Search (500) & FRAMES (824) \\
 & Tool Interaction & TMax (300) & BFCL V4 (5,106) \\
\bottomrule
\end{tabular*}
\end{table}

All models use chain-of-thought reasoning, and we report empirical pass@4 over the temperature--seed pairs $(0,42)$, $(0.2,43)$, $(0.4,44)$, and $(0.6,45)$; a task counts as solved if any attempt passes the verifier. We aggregate the benchmark scores $s_b\in[0,1]$ with the geometric mean $\mathrm{GM}=100\bigl(\prod_{b=1}^{8}s_b\bigr)^{1/8}$, which weights the four ID and four OOD benchmarks equally (task details in Appendix~\ref{app:evaluation-tasks}).

\section{Experimental Results}

Tables~\ref{tab:reasoning-results} and~\ref{tab:agentic-results} report the results. SoFT achieves the highest GM in all three settings, while individual benchmarks reveal the distinct strengths of other methods.

\subsection{Main Results on Reasoning Tasks}

\textbf{Challenges of mixed-data training.} Fine-tuning gains are uneven across reasoning tasks (Table~\ref{tab:reasoning-results}). At 7B, Confidence leads on APPS and ties for the best Science score, yet its Math score falls from 52.00 to 38.67 and ARC-C from 92.11 to 75.00, leaving its GM (60.24) below that of Base (62.91). SFT likewise improves Science but loses more than nine points on both Math and MBPP+. Overall, four baselines at 7B and two at 1.5B finish below Base in GM: training on the mixture can strengthen one part of it while weakening another.

\textbf{SoFT under mixed-data training.} SoFT improves nearly every reasoning benchmark. It raises all eight scores over Base at 1.5B and seven at 7B, matching Base on the remaining task, and its GMs of 54.17 and 65.88 are the highest at both scales. At 7B, Science rises from 82.70 to 88.30 and MBPP+ from 68.78 to 75.66. DFT remains strongest on 7B Math and Confidence on 7B APPS, but SoFT's broader gains yield the highest aggregate score.

\begin{table}[!t]
\vspace{-0.1in}
\centering
\caption{Reasoning performance (\%) under the fixed four-attempt schedule. Bold marks column bests per model size, including ties. Math: OpenR1-Math; Riddle: RiddleSense.}
\label{tab:reasoning-results}
\vspace{0.07in}
\begingroup
\setlength{\tabcolsep}{4pt}
\fontencoding{T1}\selectfont
\renewcommand{\arraystretch}{1.08}
\resizebox{\linewidth}{!}{%
\begin{tabular}{@{}cl@{\hspace{9pt}}rrrr@{\hspace{10pt}}rrrr@{\hspace{9pt}}r@{}}
\toprule[0.9pt]
 & & \multicolumn{4}{c}{\textbf{In-distribution (ID)}} & \multicolumn{4}{c}{\textbf{Out-of-distribution (OOD)}} & \\[2pt]
Model & Method & Math & APPS & Riddle & Science & GSM8K & ARC-C & BoolQ & MBPP+ & \textbf{GM} \\
\midrule[0.7pt]
\multirow{8}{*}{\rotatebox{90}{Qwen2.5-1.5B}} & Base & 34.00 & 6.80 & 66.50 & 76.50 & 87.34 & 85.03 & 82.91 & 60.85 & 50.76 \\
 & SFT & 37.00 & \textbf{8.00} & 70.73 & 78.30 & 87.11 & 88.49 & 82.77 & 63.76 & 53.46 \\
 & Confidence & 33.33 & 6.00 & 71.81 & 79.00 & 85.06 & \textbf{88.82} & 83.05 & 62.70 & 50.84 \\
 & Rho-1 & 35.00 & 6.80 & 72.00 & 77.30 & 86.05 & 87.99 & 83.47 & 60.58 & 51.66 \\
 & DFT & 31.00 & 6.40 & 68.66 & 73.40 & 83.55 & 86.35 & 81.21 & 60.58 & 49.40 \\
 & PriFT-mass & 36.00 & 5.60 & 70.04 & 75.60 & 85.90 & 83.39 & 84.18 & 65.61 & 50.49 \\
 & L2-SP & 35.33 & 6.80 & 70.53 & 79.20 & 86.28 & \textbf{88.82} & 82.91 & 65.87 & 52.32 \\
 & \textbf{SoFT} & \textbf{37.33} & \textbf{8.00} & \textbf{72.30} & \textbf{79.80} & \textbf{87.49} & 86.18 & \textbf{85.59} & \textbf{66.67} & \textbf{54.17} \\
\midrule[0.35pt]
\multirow{8}{*}{\rotatebox{90}{Qwen2.5-7B}} & Base & 52.00 & 12.80 & 82.61 & 82.70 & 91.74 & 92.11 & 92.80 & 68.78 & 62.91 \\
 & SFT & 42.33 & 11.60 & 84.28 & 87.10 & 94.92 & 82.40 & 93.50 & 59.26 & 59.46 \\
 & Confidence & 38.67 & \textbf{13.20} & 84.87 & \textbf{88.30} & 94.62 & 75.00 & 90.82 & 70.37 & 60.24 \\
 & Rho-1 & 24.33 & 10.40 & 76.72 & 85.60 & 89.39 & 79.77 & 83.47 & 30.16 & 48.34 \\
 & DFT & \textbf{58.00} & 11.20 & 82.81 & 82.70 & 93.71 & 93.91 & 90.96 & 71.69 & 63.22 \\
 & PriFT-mass & 55.33 & 12.80 & 83.30 & 83.60 & 94.69 & 95.23 & 92.23 & 74.34 & 64.64 \\
 & L2-SP & 43.00 & 12.80 & 84.09 & 84.80 & 94.24 & 96.05 & 93.64 & 72.49 & 62.77 \\
 & \textbf{SoFT} & 55.33 & 12.80 & \textbf{85.56} & \textbf{88.30} & \textbf{95.38} & \textbf{97.37} & \textbf{94.35} & \textbf{75.66} & \textbf{65.88} \\
\bottomrule[0.9pt]
\end{tabular}}
\endgroup
\end{table}

\subsection{Results on Agentic Tasks}

\textbf{Challenges of mixed-data training.} Agentic tasks show a similar imbalance (Table~\ref{tab:agentic-results}). SFT leads on TMax and BigCodeBench, yet its $\tau^2$ ID score falls from 48.33 to 35.00 and its BFCL V4 score from 40.20 to 21.10. Confidence and L2-SP also finish below Base in GM, so coding and terminal gains can come at the cost of customer interaction and tool calling.

\begin{table}[!t]
\centering
\caption{Agentic performance (\%) under the fixed four-attempt schedule. Bold marks column bests, including ties. LCB: LiveCodeBench; BCB: BigCodeBench.}
\label{tab:agentic-results}
\vspace{0.07in}
\begingroup
\setlength{\tabcolsep}{4pt}
\fontencoding{T1}\selectfont
\renewcommand{\arraystretch}{1.10}
\resizebox{\linewidth}{!}{%
\begin{tabular}{@{}cl@{\hspace{9pt}}rrrr@{\hspace{10pt}}rrrr@{\hspace{9pt}}r@{}}
\toprule[0.9pt]
 & & \multicolumn{4}{c}{\textbf{In-distribution (ID)}} & \multicolumn{4}{c}{\textbf{Out-of-distribution (OOD)}} & \\[2pt]
Model & Method & $\tau^2$ & LCB & Search & TMax & $\tau^2$ & BCB & BFCL V4 & FRAMES & \textbf{GM} \\
\midrule[0.7pt]
\multirow{8}{*}{\rotatebox{90}{Qwen3-8B}} & Base & 48.33 & 39.05 & 23.80 & 7.33 & 20.67 & 53.51 & 40.20 & 46.12 & 30.11 \\
 & SFT & 35.00 & 37.32 & 26.20 & \textbf{16.00} & 32.00 & \textbf{58.20} & 21.10 & 46.00 & 31.58 \\
 & Confidence & 40.00 & 35.59 & 17.60 & 5.00 & 22.00 & 56.32 & 34.15 & 48.67 & 26.70 \\
 & Rho-1 & 58.33 & \textbf{40.20} & 22.80 & 9.00 & 29.33 & 53.51 & 40.60 & 45.15 & 32.93 \\
 & DFT & 56.67 & 37.32 & 23.00 & 5.33 & 27.67 & 52.02 & 41.32 & 44.66 & 30.18 \\
 & PriFT-mass & 56.67 & 39.48 & 20.40 & 7.67 & 26.00 & 52.89 & 40.02 & 43.93 & 30.97 \\
 & L2-SP & 26.67 & 37.90 & 18.20 & 7.67 & 29.00 & 55.57 & 33.92 & 48.54 & 27.97 \\
 & \textbf{SoFT} & \textbf{65.00} & 38.04 & \textbf{27.20} & 9.33 & \textbf{48.33} & 56.52 & \textbf{41.36} & \textbf{48.79} & \textbf{36.93} \\
\bottomrule[0.9pt]
\end{tabular}}
\vspace{-0.1in}
\endgroup
\end{table}

\textbf{SoFT under mixed-data training.} SoFT improves seven of eight agentic scores over Base and achieves the highest GM, 36.93. Its largest gains come from $\tau^2$: ID rises from 48.33 to 65.00 and OOD from 20.67 to 48.33. The only regression is a slight dip on LiveCodeBench, from 39.05 to 38.04, while SFT still leads on TMax and BigCodeBench.

\subsection{Shared Patterns across Task Families}

Across task families, the key difference is not whether a method achieves a high score on one benchmark, but whether it does so without large losses elsewhere. For example, 8B SFT raises the four-benchmark ID geometric mean from 23.95 to 27.20 but lowers the OOD mean from 37.84 to 36.67.

SoFT instead raises both means in every setting: to 28.14 and 48.45 at 8B, from 32.93 to 36.23 and 78.24 to 80.99 at 1.5B, and from 46.18 to 48.10 and 85.70 to 90.23 at 7B. Overall, it improves 22 of 24 scores over Base, matches one, and falls slightly below on one.

\section{Exploratory Analysis}
\label{sec:exploration}

We examine three consequences of SoFT's coupled objective (Eq.~\ref{eq:soft-loss}). \textbf{RQ1} relates benchmark gains to policy drift from Base. \textbf{RQ2} measures how much probability the trained model assigns to demonstrated tokens. \textbf{RQ3} examines how $R^\star$ distributes target changes across domains.

\subsection{RQ1: Performance Gain versus Policy Drift}
\label{sec:explore-preservation}

\begin{figure}[!t]
\centering
\includegraphics[width=0.8\linewidth,trim=3pt 3pt 3pt 3pt,clip]{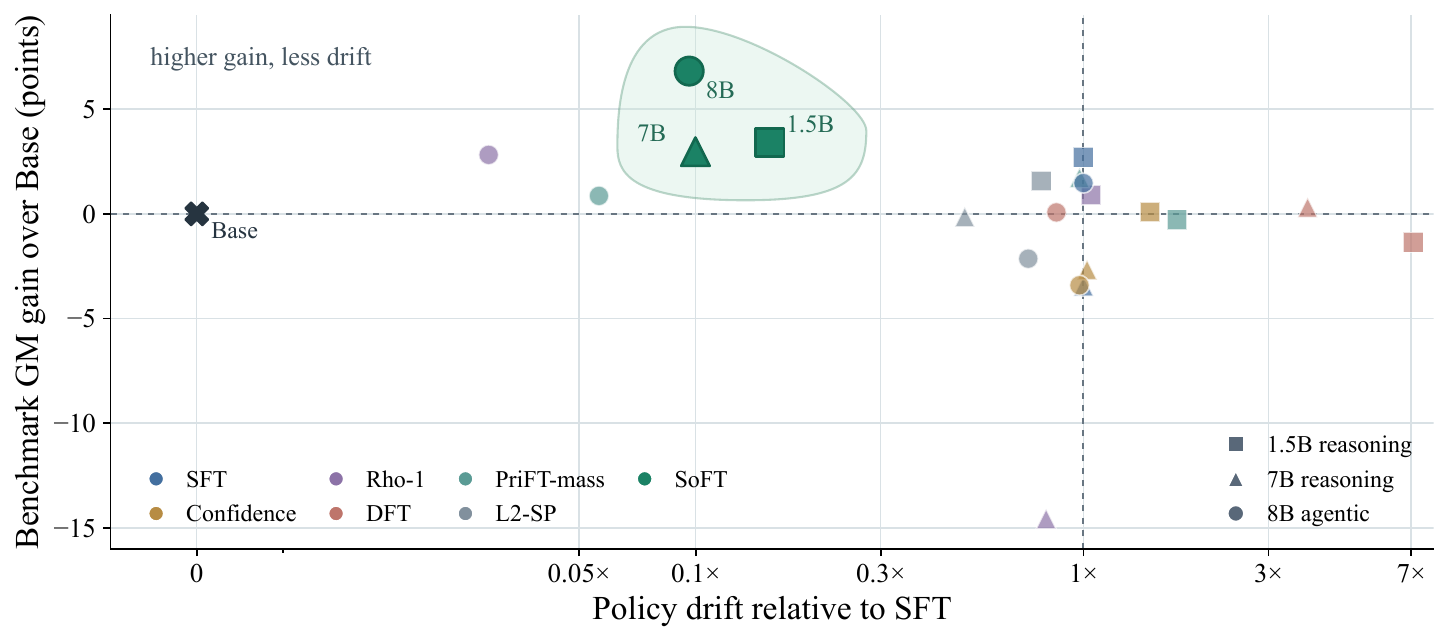}
\vspace{-0.1in}
\caption{Benchmark GM gain versus policy drift relative to SFT. The pale rounded region highlights the three SoFT points. Colors denote methods, shapes denote settings, and the horizontal axis uses a symmetric log scale near zero.}
\label{fig:base-retention}
\end{figure}

\textbf{SoFT gains more with less policy drift.} Figure~\ref{fig:base-retention} compares the GM gain of each trained model with its policy drift from Base. We measure drift as $\mathrm{KL}(\pi_0\|\pi_\theta)$ on held-out teacher-forced prefixes and divide by SFT's KL within the same setting, so SFT lies at one on the horizontal axis. SoFT occupies the favorable upper-left region at all three scales. Its relative drift is about one tenth of SFT's, and its GM gain is the largest in each setting, peaking at 8B.

\begin{figure}[!t]
\centering
\includegraphics[width=\linewidth,trim=3pt 3pt 3pt 3pt,clip]{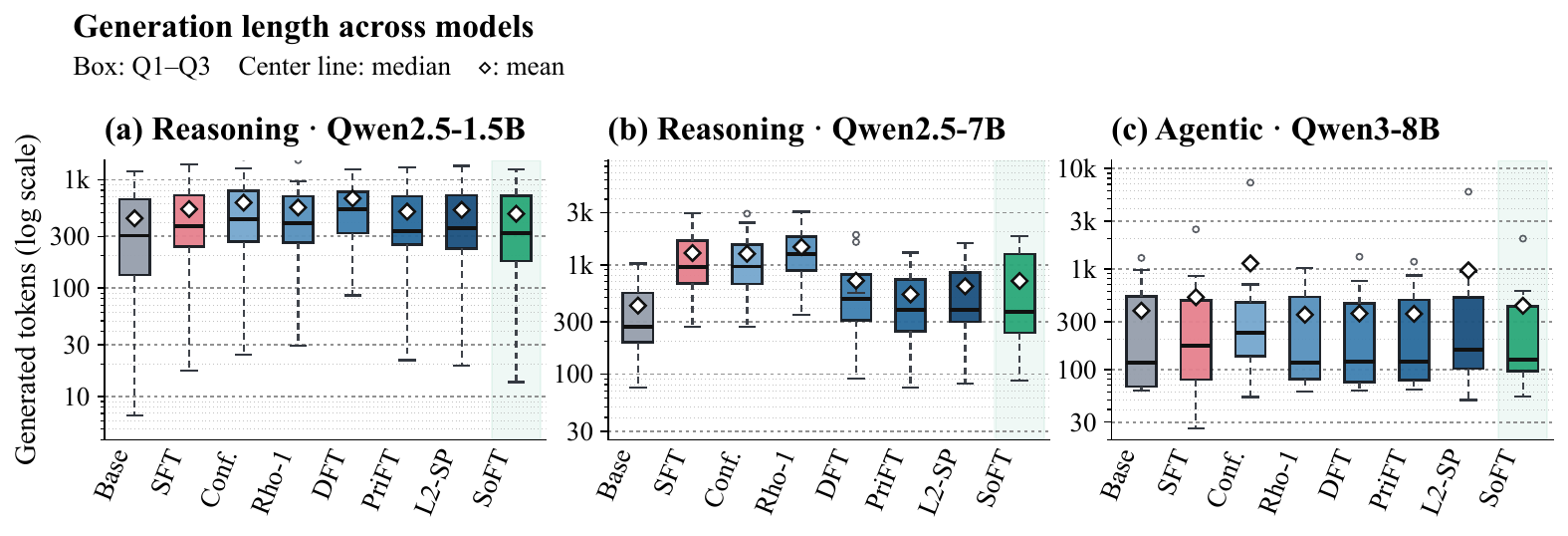}
\vspace{-0.30in}
\caption{First-attempt output lengths for all eight methods across eight benchmark means per setting. Boxes: Q1--Q3; center lines: medians; diamonds: means. Agentic lengths cover scorer-facing outputs, not total rollout cost.}
\label{fig:generation-length}
\vspace{-0.1in}
\end{figure}

\textbf{Output similarity and task success capture different forms of preservation.} On reasoning tasks, SoFT's output-token frequencies stay closer to Base than SFT's at both scales (mean Jensen--Shannon divergence: 0.044 versus 0.090 at 1.5B; 0.065 versus 0.169 at 7B). SoFT's response length is also closer to Base than SFT's on 15 of 16 benchmarks (Figure~\ref{fig:generation-length}). L2-SP offers a useful counterexample at 7B: its token frequencies are closer to Base, yet it retains fewer Base successes. Lexical similarity and task success thus describe different aspects of preservation. Appendix~\ref{app:policy-drift} reports the absolute KL values and a paired retention analysis.

\subsection{RQ2: Realized Demonstration Learning}
\label{sec:explore-acquisition}

\begin{figure}[!t]
\centering
\includegraphics[width=\linewidth,trim=1pt 3pt 1pt 2pt,clip]{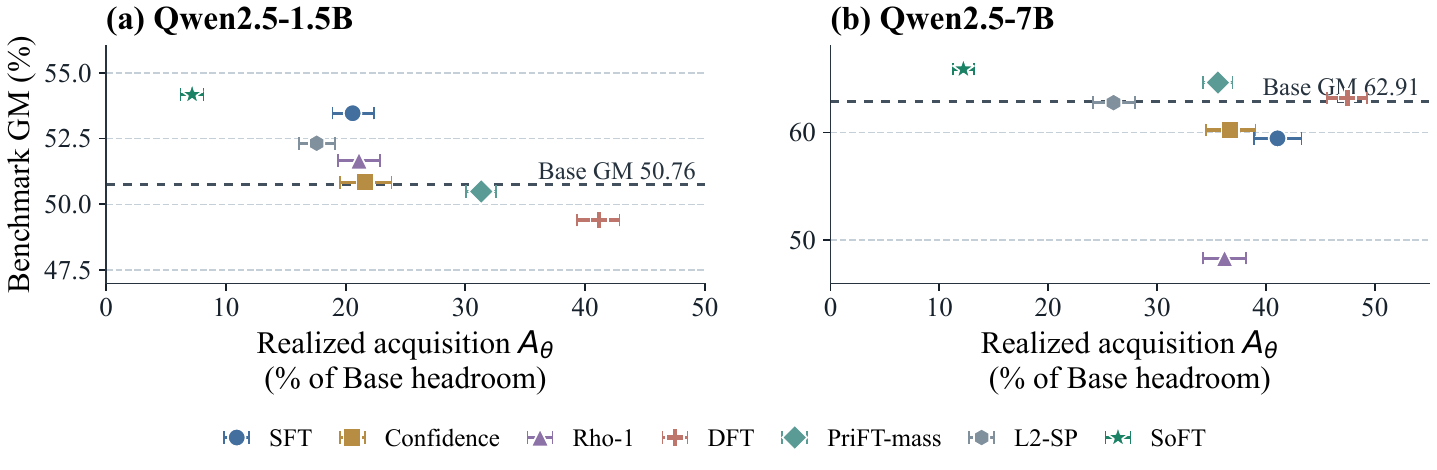}
\vspace{-0.30in}
\caption{Realized acquisition $A_\theta$ versus benchmark GM for seven methods at each reasoning scale. Horizontal bars show exploratory resampling ranges; dashed lines mark Base GM. The validation subset was used during hyperparameter selection.}
\label{fig:acquisition-performance}
\vspace{-0.1in}
\end{figure}

\textbf{More demonstration uptake does not guarantee better performance.} We measure how training changes the probability of demonstrated tokens at shared teacher-forced states. We define $A_\theta=\sum_t[\pi_\theta(y_t^\star\mid s_t)-\pi_0(y_t^\star\mid s_t)]/\sum_t[1-\pi_0(y_t^\star\mid s_t)]$. It measures the share of Base's remaining gold-token probability (Base headroom) acquired by the trained model; on SoFT's constructed target, it equals $R_{d(x)}^\star$. Figure~\ref{fig:acquisition-performance} computes $A_\theta$ on the first 128 tokens of 32 validation trajectories per reasoning domain. At 1.5B, SoFT has the smallest $A_\theta$ (7.2\%) and the highest GM (54.17), while DFT has the largest $A_\theta$ (41.2\%) and a lower GM (49.40). At 7B, PriFT-mass, Confidence, and Rho-1 have acquisition ratios near 36\%, while their GMs span 48.34 to 64.64. A larger $A_\theta$ thus does not imply better task performance, and SoFT's high GM with a small $A_\theta$ is consistent with preserving Base-supported alternatives. Appendix~\ref{app:realized-acquisition} gives per-domain results.

\subsection{RQ3: Effect of the Gradient Budget}
\label{sec:explore-threshold}

\begin{figure}[!t]
\centering
\includegraphics[width=0.87\linewidth,trim=3pt 4pt 3pt 4pt,clip]{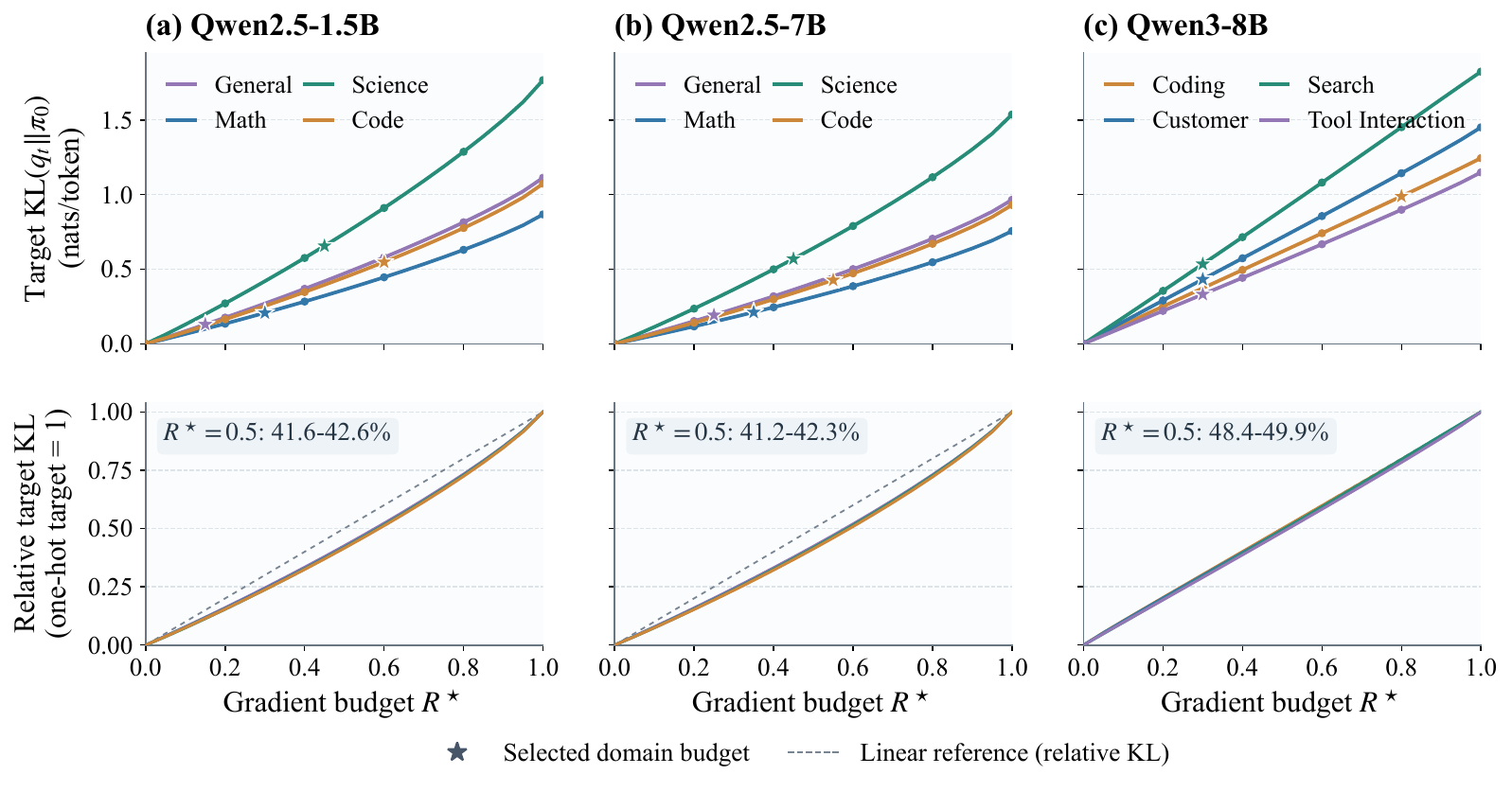}
\vspace{-0.2in}
\caption{Target KL from Base as the gradient budget varies. Top: absolute KL of the constructed target; bottom: KL relative to each trajectory's one-hot endpoint. Stars mark selected budgets.}
\label{fig:rstar-target-kl-main}
\end{figure}

\textbf{A shared budget produces comparable relative adjustments.} We vary $R^\star$ and measure $\mathrm{KL}(q_t\|\pi_0)$ for the resulting targets. In Figure~\ref{fig:rstar-target-kl-main}, Science requires a larger absolute change than Math at $R^\star=0.5$ on 1.5B (0.739 versus 0.363 nats per token). The lower panels normalize each trajectory's KL by its value at the one-hot endpoint. At the same budget, the domain averages cluster within 41.6--42.6\% at 1.5B and 48.4--49.9\% across the four agentic groups at 8B. Domains with different absolute target changes thus use similar fractions of their available change, making $R^\star$ an interpretable control across the mixture.

\textbf{The same budget activates different fractions of tokens.} At $R^\star=0.3$, positive demonstration weights apply to 16.9\% of Search tokens and 19.8\% of Customer tokens. Even within Tool Interaction, the Terminal and Tool Calling sources differ more sharply (9.3\% versus 25.5\%). Tokens that Base already predicts above the sequence floor keep their Base target, while tokens with lower Base probability receive more weight. As the budget rises, the solved floor activates additional positions rather than raising every weight equally, letting a domain budget respond to its Base confidence profile (Appendices~\ref{app:token-allocation} and~\ref{app:normalized-target-kl}).

\textbf{Selected budgets express domain priorities.} The coding domain receives the largest budget in every setting (stars in Figure~\ref{fig:rstar-target-kl-main}); validation sets each domain's learning strength, and the sequence-specific floor adapts it to Base confidence.

\section{Conclusion}
\label{sec:conclusion}

We studied multi-teacher, multi-domain SFT, where a student must learn diverse behaviors without losing generalization. SoFT uses calibrated soft targets to couple learning from demonstrations with preservation of the Base distribution. It has the best aggregate score across reasoning and agentic tasks in all three settings, although other methods lead on some benchmarks. The analysis also shows that stronger imitation of demonstrated tokens alone does not track the best aggregate task performance. These results suggest that controlling learning and preservation together can improve mixed-data fine-tuning without teacher logits or online rollouts. Future work will test whether this helps subsequent reinforcement learning. The broader lesson is to calibrate new supervision against existing knowledge.

\newpage
\section*{AI use statement}

Language models were used as objects of study and to generate teacher trajectories for the experiments described in this paper. Beyond these experimental uses, generative AI tools were used only to assist with implementing parts of the experimental code. They were not used to draft, edit, or revise the manuscript, nor to generate the text of tables or figures. The authors reviewed all experimental code, including AI-assisted implementations, to check its correctness and consistency with the methods described in the paper. All analyses, interpretations of the experimental results, and scientific claims were developed and assessed by the authors. The authors take full responsibility for the experimental methodology, code, results, claims, and final manuscript.

\section*{Ethics statement}

This work studies how language models learn from and retain capabilities
after distillation. The experiments use published datasets and benchmark
environments, together with generated teacher trajectories; they do not
involve direct interaction with human participants. Model outputs and
published trajectories may nonetheless reflect errors, biases, or
sensitive content present in their sources. We report benchmark-specific
limitations and avoid treating aggregate scores as evidence that every
capability improves. The authors are responsible for respecting the
licenses and usage conditions of the datasets and models used in this work.

\section*{Reproducibility statement}

The main text specifies the training objectives, model families,
hyperparameter-selection procedure, and evaluation protocols.
Appendix~\ref{app:training-dataset} documents the training-data sources,
teacher provenance, split sizes, token statistics, and examples. We use
fixed data splits and seeds, shared evaluation tasks within each model-size
comparison, and fixed scoring denominators. We distinguish the four-attempt
scores, versioned answer-extraction procedures, and scale-specific
finalization settings. Upon acceptance, we will publicly release the
training and evaluation code, configurations, data manifests, and processed
datasets that we are permitted to redistribute. For third-party data that
cannot be redistributed, we will provide source links and reconstruction
scripts.

\newpage
\bibliography{iclr2027_conference}
\bibliographystyle{iclr2027_conference}

\newpage
\appendix

\section{Theoretical Analysis of SoFT}
\label{app:soft-theory}

This section derives the SoFT target as a constrained KL projection, connects it to token-weighted SFT and Base-policy regularization, distinguishes it from DFT, and interprets the gradient budget $R^\star$.

Fix the teacher-forced state $s_t=(x,y_{<t}^\star)$ used in Section~\ref{sec:soft-target}, and let $y_t^\star$ be its demonstrated next token. We retain the main-text notation $q_t$, $a_t$, and $\lambda_t=1-a_t$ throughout.

\subsection{Constrained KL projection}
\label{app:soft-projection}

SoFT sets a probability floor $\tau_x$ for the demonstrated token $y_t^\star$. Among all targets that satisfy this floor, it selects the one closest to Base:
\begin{equation}
    q_t
    =
    \argmin_{q\in\Delta(\mathcal V)}
    \mathrm{KL}\!\left(q\,\|\,\pi_0(\cdot\mid s_t)\right)
    \quad
    \mathrm{s.t.}
    \quad
    q(y_t^\star)\geq\tau_x.
    \label{eq:app-kl-projection}
\end{equation}

When $\pi_0(y_t^\star\mid s_t)\geq\tau_x$, Base already satisfies the constraint and gives $q_t=\pi_0(\cdot\mid s_t)$. When $\pi_0(y_t^\star\mid s_t)<\tau_x$, the target assigns $\tau_x$ to $y_t^\star$. The Lagrangian determines the remaining mass:
\begin{equation}
    \mathcal J(q,\eta_t)
    =
    \sum_{v\neq y_t^\star}
    q(v)\log\frac{q(v)}{\pi_0(v\mid s_t)}
    +
    \eta_t
    \left(
        \sum_{v\neq y_t^\star}q(v)-(1-\tau_x)
    \right).
\end{equation}
Stationarity gives
\begin{equation}
    \log\frac{q(v)}{\pi_0(v\mid s_t)}+1+\eta_t=0,
    \qquad v\neq y_t^\star,
\end{equation}
so all non-demonstrated probabilities must be rescaled by the same factor. Normalization then yields
\begin{equation}
    \boxed{
    \begin{gathered}
        q_t(y_t^\star)
        =
        \max\{\pi_0(y_t^\star\mid s_t),\tau_x\},
        \\[3pt]
        q_t(v)
        =
        \pi_0(v\mid s_t)
        \frac{1-q_t(y_t^\star)}
             {1-\pi_0(y_t^\star\mid s_t)},
        \qquad v\neq y_t^\star.
    \end{gathered}
    }
    \label{eq:app-soft-target}
\end{equation}

For any $u,v\neq y_t^\star$, the solution preserves $q_t(u)/q_t(v)=\pi_0(u\mid s_t)/\pi_0(v\mid s_t)$. Thus, SoFT makes the minimum KL change needed to support the demonstrated token while keeping Base's relative preferences among alternatives.

\subsection{Soft target decomposition}
\label{app:soft-decomposition}

Using the main-text coefficients,
\begin{equation}
    \begin{aligned}
    a_t
    &=
    \begin{cases}
    \dfrac{\tau_x-\pi_0(y_t^\star\mid s_t)}{1-\pi_0(y_t^\star\mid s_t)},
    & \pi_0(y_t^\star\mid s_t)<\tau_x,\\[5pt]
    0,
    & \pi_0(y_t^\star\mid s_t)\geq\tau_x,
    \end{cases}
    \qquad
    \lambda_t=1-a_t,
    \\
    q_t
    &=
    a_t\delta_{y_t^\star}
    +\lambda_t\pi_0(\cdot\mid s_t).
    \end{aligned}
    \label{eq:app-target-mixture}
\end{equation}
The mixture assigns $\max\{\pi_0(y_t^\star\mid s_t),\tau_x\}$ to $y_t^\star$ and rescales the remaining Base mass proportionally, exactly recovering Eq.~\ref{eq:app-soft-target}.

Let $\mathcal L_{\mathrm{SFT},t}=-\log\pi_\theta(y_t^\star\mid s_t)$. Using the linearity of cross-entropy in its target,
\begin{align}
    \mathcal L_{\mathrm{SoFT},t}
    &=
    \mathrm{CE}\!\left(q_t,\pi_\theta(\cdot\mid s_t)\right)\\
    &=
    a_t
    \mathrm{CE}\!\left(\delta_{y_t^\star},\pi_\theta(\cdot\mid s_t)\right)
    +\lambda_t
    \mathrm{CE}\!\left(\pi_0(\cdot\mid s_t),\pi_\theta(\cdot\mid s_t)\right)\\
    &=
    a_t\mathcal L_{\mathrm{SFT},t}
    +\lambda_t\mathrm{KL}\!\left(\pi_0(\cdot\mid s_t)\,\|\,\pi_\theta(\cdot\mid s_t)\right)
    +\lambda_t H\!\left(\pi_0(\cdot\mid s_t)\right).
\end{align}
The Base entropy is independent of $\theta$. Therefore,
\begin{equation}
    \boxed{
    \mathcal L_{\mathrm{SoFT},t}
    =
    a_t\mathcal L_{\mathrm{SFT},t}
    +\lambda_t\mathrm{KL}\!\left(\pi_0(\cdot\mid s_t)\,\|\,\pi_\theta(\cdot\mid s_t)\right)
    +\mathrm{constant}.
    }
    \label{eq:app-soft-decomposition}
\end{equation}

SoFT combines demonstrated-token learning with Base-policy preservation at each token. The probability floor determines the strength of both terms through $a_t+\lambda_t=1$.

\subsection{Relationship to DFT}
\label{app:soft-dft}

Let $z_t$ denote the student logits at state $s_t$. From Eq.~\ref{eq:app-target-mixture}, the SoFT gradient is
\begin{equation}
    \nabla_{z_t}\mathcal L_{\mathrm{SoFT},t}
    =
    a_t\!\left(\pi_\theta(\cdot\mid s_t)-\delta_{y_t^\star}\right)
    +\lambda_t\!\left(\pi_\theta(\cdot\mid s_t)-\pi_0(\cdot\mid s_t)\right).
    \label{eq:app-soft-gradient}
\end{equation}
At initialization, $\pi_\theta(\cdot\mid s_t)=\pi_0(\cdot\mid s_t)$, so
\begin{equation}
    \left.
    \nabla_{z_t}\mathcal L_{\mathrm{SoFT},t}
    \right|_{\theta=\theta_0}
    =
    a_t\!\left(\pi_0(\cdot\mid s_t)-\delta_{y_t^\star}\right).
    \label{eq:app-soft-initial-gradient}
\end{equation}
Thus, the initial SoFT update is identical to token-weighted SFT with detached weight $a_t$.

DFT rescales the one-hot SFT gradient by the current student probability of the demonstrated token~\citep{wu2026dft}:
\begin{equation}
    \begin{aligned}
        \mathcal L_{\mathrm{DFT},t}
        &=
        -\operatorname{sg}[\pi_\theta(y_t^\star\mid s_t)]
        \log\pi_\theta(y_t^\star\mid s_t),\\
        \nabla_{z_t}\mathcal L_{\mathrm{DFT},t}
        &=
        \operatorname{sg}[\pi_\theta(y_t^\star\mid s_t)]
        \left(\pi_\theta(\cdot\mid s_t)-\delta_{y_t^\star}\right).
    \end{aligned}
    \label{eq:app-dft}
\end{equation}
Here, $\operatorname{sg}$ denotes the stop-gradient operator. At initialization, DFT uses the weight $\pi_0(y_t^\star\mid s_t)$, whereas SoFT uses $a_t$.

For a closer comparison, consider weighted SFT with the fixed weight $a_t$, $\mathcal L_{\mathrm{matched},t}=-a_t\log\pi_\theta(y_t^\star\mid s_t)$, which has the same initial gradient as SoFT. As the student moves away from Base, the two gradients differ by
\begin{equation}
    \nabla_{z_t}\mathcal L_{\mathrm{SoFT},t}
    -
    \nabla_{z_t}\mathcal L_{\mathrm{matched},t}
    =
    \lambda_t\!\left(\pi_\theta(\cdot\mid s_t)-\pi_0(\cdot\mid s_t)\right).
    \label{eq:app-soft-matched-difference}
\end{equation}

DFT and weighted SFT only rescale updates toward the one-hot target $\delta_{y_t^\star}$. SoFT additionally adds a restoring gradient toward Base and is minimized at $q_t$, so it changes both the strength and the destination of the update.

\subsection{Interpretation of \texorpdfstring{$R^\star$}{R*}}
\label{app:soft-rstar}

At initialization, the gold-logit gradient magnitudes of SFT and SoFT are
\begin{equation}
    \begin{aligned}
        \left|
        \frac{\partial\mathcal L_{\mathrm{SFT},t}}
             {\partial z_{t,y_t^\star}}
        \right|
        &=
        1-\pi_0(y_t^\star\mid s_t),\\
        \left|
        \frac{\partial\mathcal L_{\mathrm{SoFT},t}}
             {\partial z_{t,y_t^\star}}
        \right|
        &=
        a_t(1-\pi_0(y_t^\star\mid s_t))
        =\max\{\tau_x-\pi_0(y_t^\star\mid s_t),0\}.
    \end{aligned}
    \label{eq:app-gradient-magnitudes}
\end{equation}

SoFT chooses the sequence-level floor $\tau_x$ such that
\begin{equation}
    \boxed{
    R^\star
    =
    \frac{
        \displaystyle
        \sum_t a_t(1-\pi_0(y_t^\star\mid s_t))
    }{
        \displaystyle
        \sum_t(1-\pi_0(y_t^\star\mid s_t))
    }
    =
    \frac{
        \displaystyle
        \sum_t\max\{\tau_x-\pi_0(y_t^\star\mid s_t),0\}
    }{
        \displaystyle
        \sum_t(1-\pi_0(y_t^\star\mid s_t))
    }.
    }
    \label{eq:app-rstar}
\end{equation}
Hence, $R^\star$ is the fraction of the initial summed SFT gold-logit update that SoFT retains. Since the non-gold gradient components are nonnegative and sum to the magnitude of the gold component, the same ratio also holds for the summed logit-gradient $\ell_1$ norms.

The right-hand side of Eq.~\ref{eq:app-rstar} is continuous, piecewise linear, and monotone in $\tau_x$. Scalar bisection therefore finds the floor efficiently. Note that the ratio describes the initial logit gradients; the ratio of full parameter gradients can change during training.

\renewcommand{\ttdefault}{lmtt}
\section{Minimal PyTorch Implementations}
\label{app:minimal-code}

Both snippets assume a standard causal-LM batch: \texttt{labels} is $-100$ at prompt and padding positions, and shifting by one position produces the teacher-forced next-token pairs.

\subsection{Standard SFT}

Standard SFT gathers the log-probability of the demonstrated token, which is equivalent to cross-entropy with the one-hot target $q_t^{\mathrm{SFT}}=\delta_{y_t^\star}$.

\begin{center}
\begin{minipage}{0.94\linewidth}
\hrule\vspace{0.4em}
\begin{verbatim}
import torch.nn.functional as F

def sft_loss(student, ids, mask, labels):
    logits = student(ids, attention_mask=mask).logits[:, :-1]
    gold = labels[:, 1:]
    keep = gold.ne(-100)
    return F.cross_entropy(logits[keep], gold[keep])
\end{verbatim}
\hrule
\end{minipage}
\end{center}

\subsection{Fixed-\texorpdfstring{$R^\star$}{R*} SoFT}
SoFT follows Eqs.~\ref{eq:floor-target}--\ref{eq:rstar}, using a frozen, evaluation-mode copy of the initial student as the reference model.

\subsubsection{Practical approximation}
\label{sec:soft-implementation}

A full-vocabulary target costs $O(BT|\mathcal V|)$ memory. We retain the Base top-$K$ tokens and the demonstrated token at each supervised position, then merge the remaining vocabulary into one tail bucket. With $\mathcal S_t=\{y_t^\star\}\cup\operatorname{TopK}(\pi_0(\cdot\mid s_t))$, the loss becomes
\begin{equation}
    \widetilde{\mathcal L}_t
    = -\sum_{v\in\mathcal S_t}q_t(v)
      \log \pi_\theta(v\mid s_t)
      -q_t(\mathrm{tail})
      \log \pi_\theta(\mathrm{tail}\mid s_t).
    \label{eq:soft-topk}
\end{equation}
We use $K=32$ and include $y_t^\star$ once when it appears in the top-$K$. The approximate target keeps the exact demonstrated-token probability and the total tail mass, representing the remaining tokens as one bucket, which reduces target storage to $O(BTK)$. Caching Base probabilities avoids repeated reference forward passes, and processing student positions in chunks limits activation memory. Inference uses the student alone.

The following dense reference implementation shows the exact fixed-$R^\star$ objective before the top-$K$ approximation.

\begin{center}
\hrule\vspace{0.4em}
\begin{verbatim}
import torch
import torch.nn.functional as F

def soft_loss(student, reference, ids, mask, labels, rstar):
    logits = student(ids, attention_mask=mask).logits[:, :-1]
    gold = labels[:, 1:]
    keep = gold.ne(-100)
    gold = gold.masked_fill(~keep, 0)

    with torch.no_grad():
        ref_logits = reference(
            ids, attention_mask=mask
        ).logits[:, :-1]
        p0 = F.softmax(ref_logits, dim=-1)
        p0_gold = p0.gather(-1, gold[..., None]).squeeze(-1)

        # Solve R_x(tau_x) = R* for every sequence.
        weight = keep.to(p0.dtype)
        denom = ((1 - p0_gold) * weight).sum(-1).clamp_min(1e-12)
        lo, hi = torch.zeros_like(denom), torch.ones_like(denom)
        for _ in range(28):
            tau = (lo + hi) / 2
            numer = (((tau[:, None] - p0_gold).clamp_min(0))
                     * weight).sum(-1)
            right = numer / denom < rstar
            lo = torch.where(right, tau, lo)
            hi = torch.where(right, hi, tau)
        tau = ((lo + hi) / 2)[:, None]

        # Minimum-change target q_tau.
        q_gold = torch.maximum(p0_gold, tau)
        scale = (1 - q_gold) / (1 - p0_gold).clamp_min(1e-12)
        q = p0 * scale[..., None]
        q.scatter_(-1, gold[..., None], q_gold[..., None])

    logp = F.log_softmax(logits, dim=-1)
    token_loss = -(q * logp).sum(-1)
    return token_loss[keep].mean()
\end{verbatim}
\hrule
\end{center}

The Base probabilities define the target, the bisection loop converts the fixed budget $R^\star$ into one floor $\tau_x$ per sequence, and the last lines compute $\mathrm{CE}(q_t,\pi_\theta)$.

\section{Training Data Details}
\label{app:training-dataset}

Training uses one epoch and seed 42, with maximum sequence lengths of 8,192 for reasoning and 32,768 for agentic tasks. Methods within each setting share the data order. The selected SoFT learning rates are $10^{-5}$, $5\times10^{-6}$, and $10^{-6}$ for Qwen2.5-1.5B-Instruct, Qwen2.5-7B, and Qwen3-8B, respectively.

Counts measure training trajectories; several trajectories may share a task. Token statistics exclude validation examples. Source links identify the released datasets or environments, and teacher names follow the provenance records. We mark unknown upstream teachers as unspecified.

\subsection{Reasoning Tasks}

Both Qwen2.5 students use 36,000 training trajectories (9,000 per domain) and 800 validation examples (200 per domain). All reasoning trajectories come from public releases; Table~\ref{tab:reasoning-training-data} lists their sources and teachers. The stored \texttt{token\_length} is measured with the Qwen2.5-1.5B-Instruct tokenizer and training chat template and includes prompt, completion, and template tokens. We keep sequences of at most 8,192 tokens.

The released sources are GLM and Kimi cleaned trajectories~\citep{jackrong2026glmdata,jackrong2026kimidata}, Sky-T1~\citep{novasky2025skyt1data}, OpenThoughts-4 agreed-answer math traces~\citep{marin2026agreedmath}, Mixture-of-Thoughts~\citep{openr12025mixture}, and community GPT-OSS traces~\citep{iamboosted2026traces}. The GLM and Kimi releases derive from the original datasets by Kassadin88 and ianncity~\citep{kassadin2026glmoriginal,ianncity2026kimioriginal}. Open R1 releases Mixture-of-Thoughts~\citep{huggingface2025openr1}. Its math, code, and science sources are OpenR1-Math, CodeForces CoTs, and Llama-Nemotron~\citep{lozhkov2025openr1math,penedo2025codeforces,bercovich2025llamanemotron}. The 600 GPT-OSS trajectories use 477 prompts from OpenThoughts-114k~\citep{guha2025openthoughts} and 123 from OpenCodeReasoning~\citep{ahmad2025opencodereasoning}; GPT-OSS-20B generated the responses~\citep{openai2025gptoss}.

The manifest records 11,260 verified-correct rows, 9,000 upstream-verified rows, 13,140 cleaned rows with unverified answers, 2,000 answer-agreement rows, and 600 GPT-OSS rows without verified references (477 without a reference and 123 marked unverifiable). Verification strength therefore varies by source. Length filtering only checks that sequences fit the context window; answer-quality labels are kept separately in the manifest.

\begin{table}[!t]
\centering
\caption{Reasoning training sources, teachers, and full-sequence token statistics. Validation examples are excluded.}
\label{tab:reasoning-training-data}
\begingroup
\fontencoding{T1}\selectfont
\setlength{\tabcolsep}{4pt}
\renewcommand{\arraystretch}{1.15}
\resizebox{\linewidth}{!}{%
\begin{tabular}{lllrcrrr}
\toprule
Domain & Task-data source & Teacher & Train & Link & Min & Mean & Max \\
\midrule
Math & GLM cleaned & GLM-5.1 & 450 & \href{https://huggingface.co/datasets/Jackrong/GLM-5.1-Reasoning-1M-Cleaned}{source} & 1,371 & 5894.3 & 8,187 \\
Math & Kimi cleaned & Kimi-K2.5 & 690 & \href{https://huggingface.co/datasets/Jackrong/Kimi-K2.5-Reasoning-1M-Cleaned}{source} & 2,361 & 6561.5 & 8,191 \\
Math & Sky-T1 & QwQ-32B / STILL-2 & 2,400 & \href{https://huggingface.co/datasets/NovaSky-AI/Sky-T1_data_17k}{source} & 269 & 3927.4 & 8,165 \\
Math & OpenThoughts-4 math & Qwen3-235B-A22B & 2,000 & \href{https://huggingface.co/datasets/marin-community/open-thoughts-4-11k-math-qwen3-235b-a22b-agreed-answers}{source} & 1,535 & 6658.6 & 8,192 \\
Math & Mixture-of-Thoughts & DeepSeek-R1 & 3,460 & \href{https://huggingface.co/datasets/open-r1/Mixture-of-Thoughts}{source} & 468 & 4109.7 & 8,189 \\
Code & Sky-T1 & QwQ-32B / STILL-2 & 3,600 & \href{https://huggingface.co/datasets/NovaSky-AI/Sky-T1_data_17k}{source} & 798 & 4898.2 & 8,191 \\
Code & GPT-OSS traces & GPT-OSS-20B & 600 & \href{https://huggingface.co/datasets/iAmBoosted/gpt-oss-20b-reasoning-traces}{source} & 637 & 3264.4 & 7,978 \\
Code & Mixture-of-Thoughts & DeepSeek-R1 & 4,800 & \href{https://huggingface.co/datasets/open-r1/Mixture-of-Thoughts}{source} & 551 & 5148.1 & 8,191 \\
Science & GLM cleaned & GLM-5.1 & 3,000 & \href{https://huggingface.co/datasets/Jackrong/GLM-5.1-Reasoning-1M-Cleaned}{source} & 2,014 & 3611.5 & 8,186 \\
Science & Kimi cleaned & Kimi-K2.5 & 3,000 & \href{https://huggingface.co/datasets/Jackrong/Kimi-K2.5-Reasoning-1M-Cleaned}{source} & 1,367 & 3421.0 & 7,633 \\
Science & Mixture-of-Thoughts & DeepSeek-R1 & 3,000 & \href{https://huggingface.co/datasets/open-r1/Mixture-of-Thoughts}{source} & 389 & 2061.0 & 8,168 \\
General & GLM cleaned & GLM-5.1 & 3,000 & \href{https://huggingface.co/datasets/Jackrong/GLM-5.1-Reasoning-1M-Cleaned}{source} & 132 & 2935.9 & 8,142 \\
General & Kimi cleaned & Kimi-K2.5 & 3,000 & \href{https://huggingface.co/datasets/Jackrong/Kimi-K2.5-Reasoning-1M-Cleaned}{source} & 122 & 3399.5 & 8,180 \\
General & Sky-T1 & QwQ-32B / STILL-2 & 3,000 & \href{https://huggingface.co/datasets/NovaSky-AI/Sky-T1_data_17k}{source} & 165 & 2981.3 & 8,164 \\
\midrule
\multicolumn{3}{l}{Total} & 36,000 & -- & 122 & 3991.0 & 8,192 \\
\bottomrule
\end{tabular}}
\endgroup
\end{table}

\paragraph{Example from the training corpus.}
Below is a verified DeepSeek-R1 Math example from Mixture-of-Thoughts; the middle of its reasoning is omitted for space.
\begin{quote}
\textbf{User.} At the cross-country race, 8 runners wore white sports shirts.
4 runners wore blue sports shirts. How many runners started the
cross-country race?

\textbf{Teacher reasoning, excerpt.}
\emph{So 8 white shirts plus 4 blue shirts.}
\ldots\ \emph{8 plus 4 equals 12.} \textbf{Final answer:} $\boxed{12}$.
\end{quote}

\subsection{Agentic Tasks}

After token-level deduplication, the agentic corpus contains 8,252 training and 824 validation trajectories. Table~\ref{tab:agentic-training-data} reports the stored \texttt{sequence\_length} measured with the Qwen3-8B tokenizer and chat template. Lengths include user and system text, tool schemas, tool observations, and assistant turns. The corpus contains 53,623,153 full-sequence tokens and 7,278,365 supervised assistant tokens, with a 32,768-token limit. Teacher reasoning text is removed; actions, observations, final answers, and tool-call links are retained.

Locally collected $\tau^2$~\citep{barres2025tau2} and LiveCodeBench~\citep{jain2024livecodebench} trajectories use DeepSeek V4 Flash. We reuse published Nemotron~\citep{nvidia2025agenticdata} and OpenThoughts-Agent~\citep{raoof2026data} trajectories; the local metadata leaves their individual teachers unspecified. TMax contributes published successful trajectories~\citep{ivison2026tmax} with reasoning text removed. Final packaging removes five exact token-sequence duplicates. The 1,282 coding trajectories cover 296 unique prompts.

\begin{table}[!t]
\centering
\caption{Agentic training sources, teachers, and full-sequence token statistics. Validation examples are excluded.}
\label{tab:agentic-training-data}
\begingroup
\fontencoding{T1}\selectfont
\setlength{\tabcolsep}{4pt}
\renewcommand{\arraystretch}{1.15}
\resizebox{\linewidth}{!}{%
\begin{tabular}{lllrcrrr}
\toprule
Domain group & Task-data source & Teacher & Train & Link & Min & Mean & Max \\
\midrule
Coding & LiveCodeBench v1 & DeepSeek V4 Flash & 1,282 & \href{https://huggingface.co/datasets/livecodebench/code_generation_lite}{source} & 155 & 683.0 & 4,333 \\
\multirow{2}{*}{Customer} & Nemotron interactive & Upstream; unspecified & 1,200 & \href{https://huggingface.co/datasets/nvidia/Nemotron-SFT-Agentic-v2}{source} & 1,837 & 3972.1 & 8,534 \\
 & $\tau^2$ retail/airline & DeepSeek V4 Flash & 2,997 & \href{https://github.com/sierra-research/tau2-bench}{source} & 4,899 & 9127.8 & 22,431 \\
Search & Nemotron search & Upstream; unspecified & 600 & \href{https://huggingface.co/datasets/nvidia/Nemotron-SFT-Agentic-v2}{source} & 3,892 & 18791.6 & 32,701 \\
\multirow{3}{*}{Tool Interaction} & TMax-SFT & Qwen3.6-27B & 595 & \href{https://huggingface.co/datasets/allenai/tmax-sft}{source} & 1,661 & 5575.2 & 30,575 \\
 & OpenThoughts-Agent & Upstream; unspecified & 387 & \href{https://huggingface.co/datasets/open-thoughts/OpenThoughts-Agent-v1-SFT}{source} & 949 & 2730.1 & 17,593 \\
 & Nemotron tool calling & Upstream; unspecified & 1,191 & \href{https://huggingface.co/datasets/nvidia/Nemotron-SFT-Agentic-v2}{source} & 87 & 4178.3 & 31,182 \\
\midrule
\multicolumn{3}{l}{Total} & 8,252 & -- & 87 & 6498.2 & 32,701 \\
\bottomrule
\end{tabular}}
\endgroup
\end{table}

\paragraph{Example from the training corpus.}
The following LiveCodeBench training example uses a DeepSeek V4 Flash response. We excerpt the prompt and show the complete assistant code.
\begin{quote}
\textbf{User, excerpt.} You are given an integer N between 1 and 9,
inclusive, as input. Concatenate N copies of the digit N and print the
resulting string.

\textbf{Assistant.}
\begin{verbatim}
N = int(input())
print(str(N) * N)
\end{verbatim}
\end{quote}

\section{Evaluation Tasks and Splits}
\label{app:evaluation-tasks}

\subsection{Reasoning Tasks}

The ID suite contains four task families. OpenR1-Math contributes 300 mathematical problems. APPS contributes 250 programming problems in the benchmark's difficulty proportions: 50 introductory, 150 interview, and 50 competition tasks. Science MCQ contains 1,000 multiple-choice science questions selected from Mixture-of-Thoughts. RiddleSense contributes 1,018 multiple-choice commonsense riddles.

The OOD suite tests the same four capabilities using different question sources. GSM8K contains 1,319 grade-school mathematics word problems. MBPP+ contains 378 short Python programming tasks. ARC-Challenge contributes 608 multiple-choice science questions. BoolQ contributes 708 deduplicated yes/no questions about short passages, drawn from its labeled validation split. Table~\ref{tab:evaluation-suites} summarizes the ID--OOD pairing for each domain.

\subsection{Agentic Tasks}

The ID suite covers customer interaction, coding, search, and tool interaction. The 60 $\tau^2$ tasks are held-out retail and airline customer-service scenarios involving orders, returns, flights, and bookings. LiveCodeBench contributes 694 programming problems. Search ID contains 500 held-out Nemotron questions that call for finding and combining information. TMax contributes 300 tasks in terminal environments, where the agent works with commands and files.

The OOD suite uses different task sources or domains. Its 300 $\tau^2$ cases come from telecom customer support and banking-knowledge retrieval. BigCodeBench contributes 2,280 evaluation cases: 1,140 software-oriented coding tasks presented in both completion and instruction formats. BFCL V4 contributes 5,106 function-calling cases, including tool selection, argument construction, and multi-turn interactions. FRAMES contains 824 factual questions that require retrieving and combining evidence from multiple sources.

\section{Token-Level Allocation Diagnostics}
\label{app:token-allocation}

\begin{figure}[H]
\centering
\includegraphics[width=\linewidth]{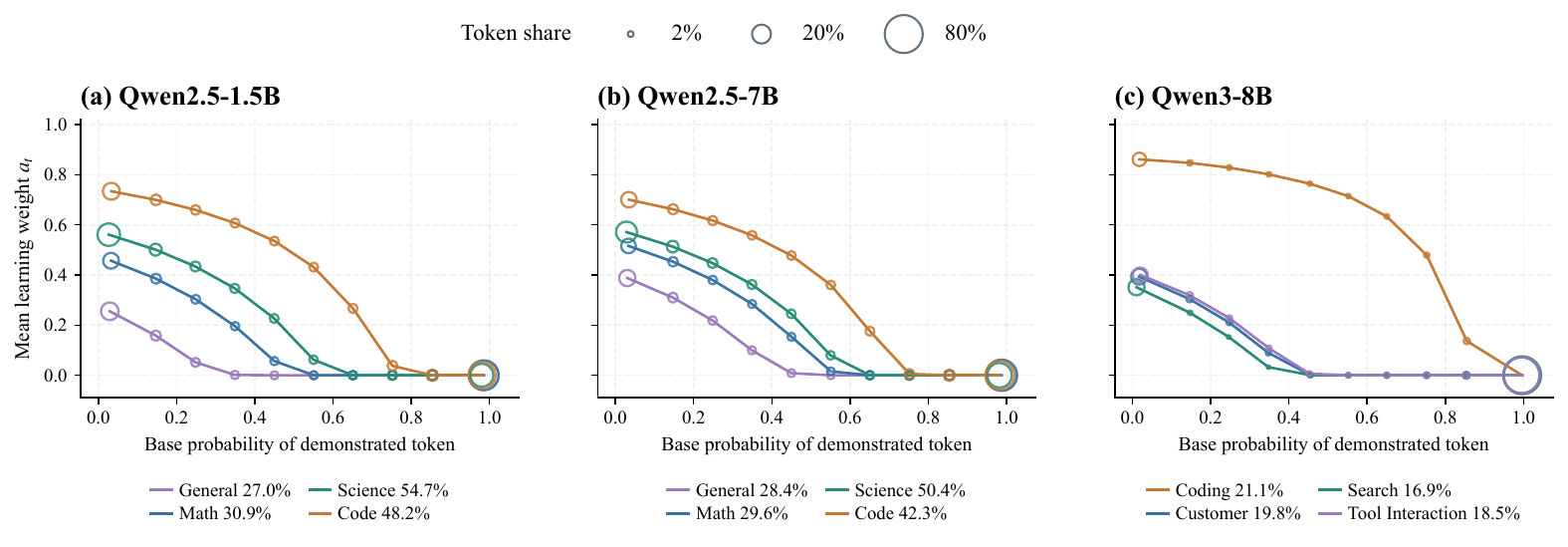}
\caption{Token-weighted target allocation from Base caches. Lines: mean $a_t$ by Base probability; circle areas: token fractions; legend percentages: fraction with $a_t>0$. The 8B panel pools source data into four domain groups.}
\label{fig:soft-learning-allocation}
\end{figure}

Figure~\ref{fig:soft-learning-allocation} reconstructs token weights from saved Base probabilities and selected domain budgets for all 36,000 trajectories at each reasoning scale and 8,252 agentic trajectories. The 8B sources are pooled into four domain groups by supervised-token counts. Weights decrease with Base confidence by construction. The empirical question is where the training data place tokens along this curve.

The adjustment is concentrated in a subset of tokens. Within 8B Tool Interaction, the Terminal source has $a_t>0$ on 9.3\% of tokens, while 87.0\% already receive Base probability above 0.9. Tokens with $a_t=0$ keep the Base distribution as their training target. The fraction of adjusted tokens also does not simply follow the budget: at 1.5B, Science uses $R^\star=0.45$ and adjusts 54.7\% of tokens, whereas Code uses $R^\star=0.60$ and adjusts 48.2\%. The budget controls the initial gold-logit gradient ratio, whereas the adjusted-token fraction depends on Base confidence.

The same budget can likewise produce different adjustment rates. The three non-coding agentic groups all use $R^\star=0.3$, yet their rates range from 16.9\% on Search to 19.8\% on Customer; Tool Interaction lies between them at 18.5\%. Its Terminal and Tool Calling sources differ more sharply, at 9.3\% and 25.5\%. Thus, the distribution of Base confidence shapes how each group and source receives supervision. Together, these measurements show how SoFT adapts token-level learning within a mixed training corpus.

\section{Policy Drift and Behavioral Retention}
\label{app:policy-drift}

Figure~\ref{fig:base-retention} normalizes each method's forward KL $\mathrm{KL}(\pi_0\|\pi_\theta)$ by that of SFT in the same setting. The reasoning measurements use the same 32 held-out validation trajectories per domain as RQ2 and average over the first 128 demonstrated tokens. The 8B measurement uses 16 validation trajectories from each of five agentic source domains and the first 128 supervised assistant tokens. Normalization places all settings on a common axis, although absolute KL still depends on the data.

\begin{figure}[!t]
\centering
\includegraphics[width=\linewidth]{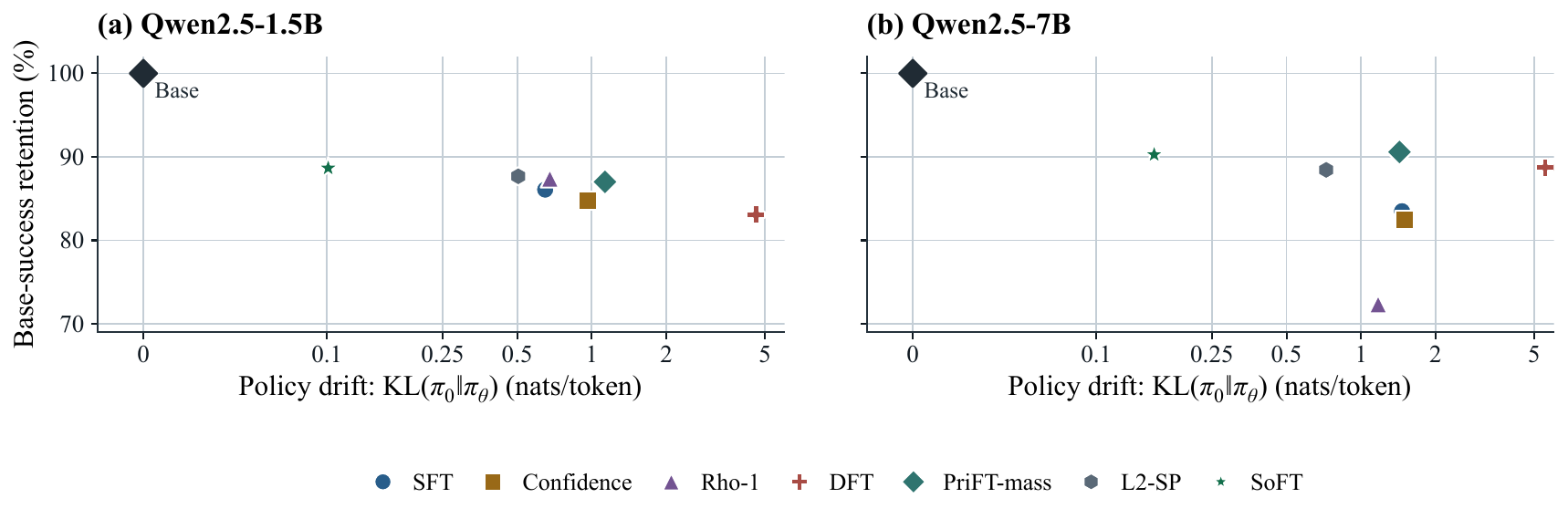}
\caption{Final policy drift versus Base-success retention on reasoning tasks. Drift is $\mathrm{KL}(\pi_0\|\pi_\theta)$ over held-out teacher-forced prefixes; retention is the equal-weight mean over eight benchmarks.}
\label{fig:policy-kl-retention}
\end{figure}

Figure~\ref{fig:policy-kl-retention} compares absolute reasoning KL with mean Base-success retention. SoFT has the lowest drift and among the highest retention at both scales. At 1.5B, drift and retention broadly move in opposite directions. At 7B, DFT moves far from Base while retaining many successes, whereas Rho-1 moves less and retains fewer. Global KL captures one aspect of preservation; task success reveals another. The policy diagnostic uses validation prefixes, while retention is measured on the separate benchmark tasks.

\section{Why Normalized Target-KL Curves Can Align}
\label{app:normalized-target-kl}

Section~\ref{sec:explore-threshold} shows nearly aligned normalized curves alongside different absolute target KL values. We use the KL chain rule, a mixture decomposition, and a low-probability limit to explain how this pattern can arise from the constructed targets.

\subsection{Reducing the full-vocabulary KL to a scalar}

Write $r\in[0,1]$ for a candidate value of $R^\star$. For trajectory $x$, let $T_x$ be the number of supervised positions. Assume $\pi_0(y_t^\star\mid s_t)>0$ and $\sum_t(1-\pi_0(y_t^\star\mid s_t))>0$. The floor $\tau_x(r)$ satisfies
\begin{equation}
\sum_t\max\{\tau_x(r)-\pi_0(y_t^\star\mid s_t),0\}
=r\sum_t(1-\pi_0(y_t^\star\mid s_t)).
\label{eq:norm-budget}
\end{equation}
At the endpoints, use the Base target for $r=0$ and the one-hot target for $r=1$. Define the binary KL and binary entropy, using natural logarithms, as
\begin{equation}
\begin{aligned}
b(u\|p)&=u\log\frac{u}{p}+(1-u)\log\frac{1-u}{1-p},\\
h(u)&=-u\log u-(1-u)\log(1-u),
\end{aligned}
\label{eq:norm-binary-defs}
\end{equation}
with the usual continuous endpoint conventions and $b(1\|1)=0$.

\paragraph{Proposition 1 (binary reduction).}
For the SoFT target, written $q_{\tau_x,t}$ in this section to make its dependence on the floor explicit,
\begin{equation}
\mathrm{KL}\!\left(q_{\tau_x,t}\,\|\,\pi_0(\cdot\mid s_t)\right)
=b\!\left(\max\{\pi_0(y_t^\star\mid s_t),\tau_x\}\,\|\,\pi_0(y_t^\star\mid s_t)\right).
\label{eq:norm-binary-reduction}
\end{equation}
\emph{Proof.} Partition the vocabulary into the demonstrated token and its complement. The KL chain rule separates the divergence into the binary mass-allocation term and a conditional KL over other tokens. SoFT preserves the Base distribution conditional on that complement, so the conditional KL is zero. When $\pi_0(y_t^\star\mid s_t)=1$, the target remains unchanged and both sides are zero.

The per-trajectory absolute and normalized changes are therefore
\begin{equation}
\begin{aligned}
D_x(r)&=\frac{1}{T_x}\sum_t b\!\left(\max\{\pi_0(y_t^\star\mid s_t),\tau_x(r)\}\,\|\,\pi_0(y_t^\star\mid s_t)\right),\\
C_x(r)&=\frac{D_x(r)}{D_x(1)},
\qquad D_x(1)=\frac{1}{T_x}\sum_t-\log\pi_0(y_t^\star\mid s_t).
\end{aligned}
\label{eq:norm-trajectory-curve}
\end{equation}
We form each normalized domain curve by averaging $C_x(r)$ over trajectories. The endpoints satisfy $C_x(0)=0$ and $C_x(1)=1$ by construction. The intermediate shape depends on the trajectories' Base probabilities and can vary across domains.

\subsection{An exact cancellation of token proportions}

Let $P$ be the Base probability of the demonstrated token at a uniformly chosen position in a trajectory. Consider the idealized mixture
\begin{equation}
P\sim(1-f)\delta_1+fF,
\qquad 0<f\leq1,
\label{eq:norm-mixture}
\end{equation}
where $\delta_1$ is a point mass at probability one and $F$ is a distribution on $(0,1)$ with finite, positive $\mathbb E_F[-\log P]$. The first component represents tokens with full Base confidence; the second contains the remaining tokens. The point mass at one makes the cancellation explicit; in finite-logit models, high-confidence probabilities only approach one.

\paragraph{Proposition 2 (mixture-proportion invariance).}
For fixed $F$, the entire normalized curve is independent of $f$.

\emph{Proof.} Both sides of the budget equation contain the same factor $f$:
\begin{equation}
f\,\mathbb E_F[(\tau-P)_+]
=rf\,\mathbb E_F[1-P].
\end{equation}
After canceling $f$, the floor depends only on $F$ and $r$. Denote it by $\tau_F(r)$. The certain tokens contribute zero KL, so
\begin{equation}
\begin{aligned}
D_x(r)&=f\,\mathbb E_F[b(\max\{P,\tau_F(r)\}\|P)],\\
D_x(1)&=f\,\mathbb E_F[-\log P],\\
C_x(r)&=\frac{\mathbb E_F[b(\max\{P,\tau_F(r)\}\|P)]}
{\mathbb E_F[-\log P]}.
\end{aligned}
\label{eq:norm-mixture-cancellation}
\end{equation}
The final expression contains no $f$.

For example, increasing the proportion of uncertain tokens from 10\% to 50\% multiplies absolute KL by five while leaving the normalized curve unchanged when $F$ stays fixed. Domains can therefore require different amounts of adjustment yet have identical normalized responses. The result extends to domain averages when they share the same distribution of trajectory-level $F$ values, even if their $f$ values differ. The full vocabulary distributions may still differ.

\subsection{A closed form and a low-probability limit}

For a simple special case, set $F=\delta_\epsilon$ with $0<\epsilon<1$. All demonstrated tokens whose Base probability is below one then have probability $\epsilon$. Equation~\ref{eq:norm-budget} gives
\begin{equation}
\tau_\epsilon(r)=\epsilon+(1-\epsilon)r,
\qquad
C_\epsilon(r)=\frac{b(\epsilon+(1-\epsilon)r\|\epsilon)}{\log(1/\epsilon)}.
\label{eq:norm-homogeneous}
\end{equation}
The proportion $f$ cancels, while the confidence level $\epsilon$ continues to shape the curve.

\paragraph{Proposition 3 (entropy bound).}
For this special case and every $r\in[0,1]$,
\begin{equation}
\max\!\left\{0,r-\frac{\log2}{\log(1/\epsilon)}\right\}
\leq C_\epsilon(r)\leq r.
\label{eq:norm-entropy-bound}
\end{equation}
Consequently, $C_\epsilon(r)$ converges uniformly to $r$ as $\epsilon\to0$.

\emph{Proof.} Put $u=\epsilon+(1-\epsilon)r$ and $L=\log(1/\epsilon)$. Expanding the binary KL gives the exact identity
\begin{equation}
C_\epsilon(r)
=u-\frac{h(u)}{L}
+\frac{(1-u)\log(1/(1-\epsilon))}{L}.
\label{eq:norm-entropy-identity}
\end{equation}
Since $u\geq r$, $h(u)\leq\log2$, and the last term is nonnegative, this yields the lower bound, together with nonnegativity of KL. For the upper bound, $\operatorname{Bern}(u)=(1-r)\operatorname{Bern}(\epsilon)+r\delta_1$. Convexity of KL in its first argument gives $b(u\|\epsilon)\leq r\log(1/\epsilon)$. The resulting uniform error bound is $0\leq r-C_\epsilon(r)\leq\log2/L$.

For sufficiently small $\epsilon$, the leading term approaches $r$. The entropy term places the curve below the diagonal. The approximation remains sensitive to confidence: at $r=0.5$, $C_\epsilon(r)$ is about 0.189 for $\epsilon=0.5$ and 0.356 for $\epsilon=0.01$. Thus, curve alignment also depends on the probability profile.

\subsection{Nearly certain tokens}

For a more realistic mixture, replace $\delta_1$ by a distribution $G$ supported on $[1-\delta,1]$. Let $F$ be supported on $(0,\epsilon]$ and suppose the floor lies between $\epsilon$ and $1-\delta$. Then all $F$ tokens are active and all $G$ tokens are inactive. Define
\begin{equation}
\begin{gathered}
\kappa=\frac{1-f}{f},\qquad \mu=\mathbb E_F[P],\qquad \nu=\mathbb E_G[1-P],\\
\ell_F=\mathbb E_F[-\log P],\quad
c_F=\mathbb E_F[-\log(1-P)],\quad
\ell_G=\mathbb E_G[-\log P].
\end{gathered}
\end{equation}
The budget equation and normalized divergence now give
\begin{equation}
\begin{aligned}
\tau(r)&=r+(1-r)\mu+r\kappa\nu,\\
C_x(r)&=\frac{\tau(r)\ell_F-h(\tau(r))+(1-\tau(r))c_F}
{\ell_F+\kappa\ell_G}.
\end{aligned}
\label{eq:norm-near-certain}
\end{equation}
Substituting the mixture into the budget equation and expanding binary KL over the active component gives these identities. The high-confidence component contributes residual gradient mass through $\kappa\nu$ and endpoint KL through $\kappa\ell_G$, restoring dependence on $f$. For fixed $F$ and $f$, the idealized result emerges as $G$ concentrates at one under the stated active-set condition. When $\kappa$ is large, even small residuals from many nearly certain tokens can affect the curve.

\subsection{A counterexample and implications for the observation}

The homogeneous case has a specific bound. To see how a mixed probability profile differs, consider equal numbers of tokens with probabilities $\epsilon$ and $1/2$, and set $r=1/2$. Both groups are active at
\begin{equation}
\tau=\frac58+\frac{\epsilon}{4}.
\end{equation}
As $\epsilon\to0$, the low-probability group dominates both KL expressions, giving
\begin{equation}
C_x(1/2)
=\frac{b(\tau\|\epsilon)+b(\tau\|1/2)}
{\log(1/\epsilon)+\log2}
\longrightarrow\frac58.
\label{eq:norm-counterexample}
\end{equation}
In contrast, the homogeneous low-probability case in Proposition 3 approaches $1/2$. Here, the normalized KL exceeds the budget, showing how different probability profiles can separate the curves.

The alignment in Section~\ref{sec:explore-threshold} is consistent with domains sharing a similar conditional probability profile while differing in the amount of required adjustment. The derivation identifies those profiles and their KL contributions as the quantities that shape the normalized curves.

\section{Training Dynamics}
\label{app:training-dynamics}

The following curves cover the 21 trained runs in the main comparisons. Reasoning and agentic runs reach 2,250 and 258 steps, respectively; Base contributes only to evaluation. We average logged training metrics in non-overlapping 50-step and 10-step windows. Held-out entropy and learning rates use their recorded points.

\subsection{Training Objectives and Predictive Entropy}

\begin{figure}[!t]
\centering
\includegraphics[width=\linewidth]{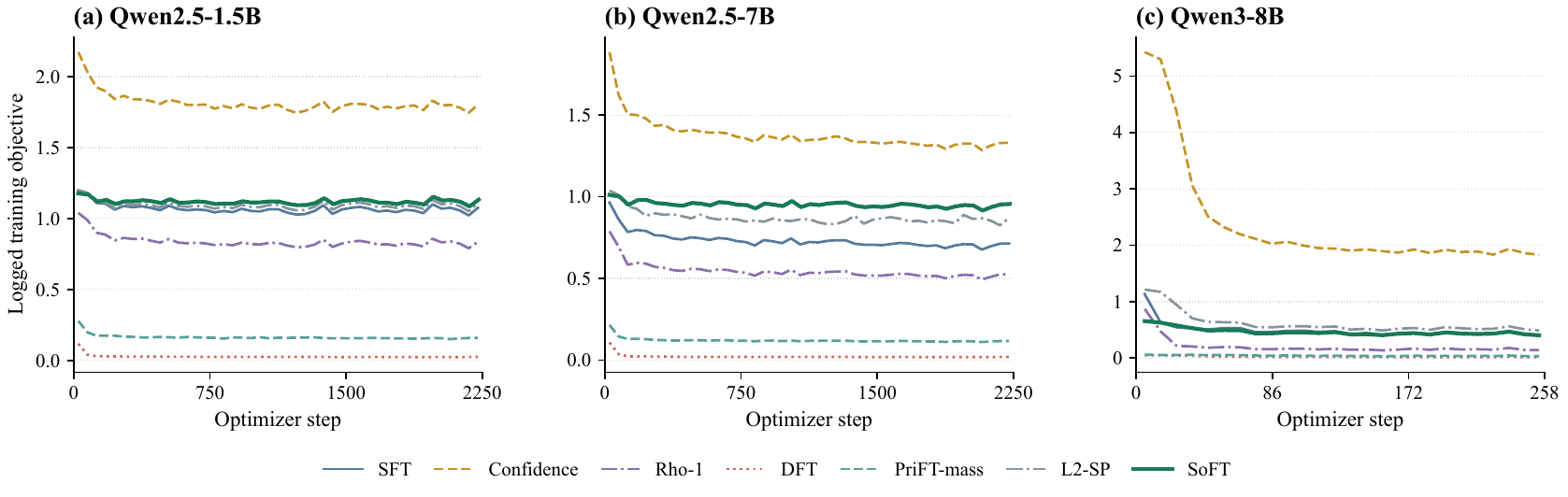}
\caption{Training objectives averaged over 50-step (reasoning) or 10-step (agentic) windows. Objective scales vary across methods; benchmark scores report task performance.}
\label{fig:training-loss}
\end{figure}

Figure~\ref{fig:training-loss} shows distinct objective scales. Weighted losses are small for DFT and PriFT-mass, while Confidence reports larger values. SoFT's reasoning loss changes little because it includes the entropy of the soft target as a constant offset. The curves describe optimization within each run; benchmark results provide the performance comparison.

\begin{figure}[!t]
\centering
\includegraphics[width=\linewidth]{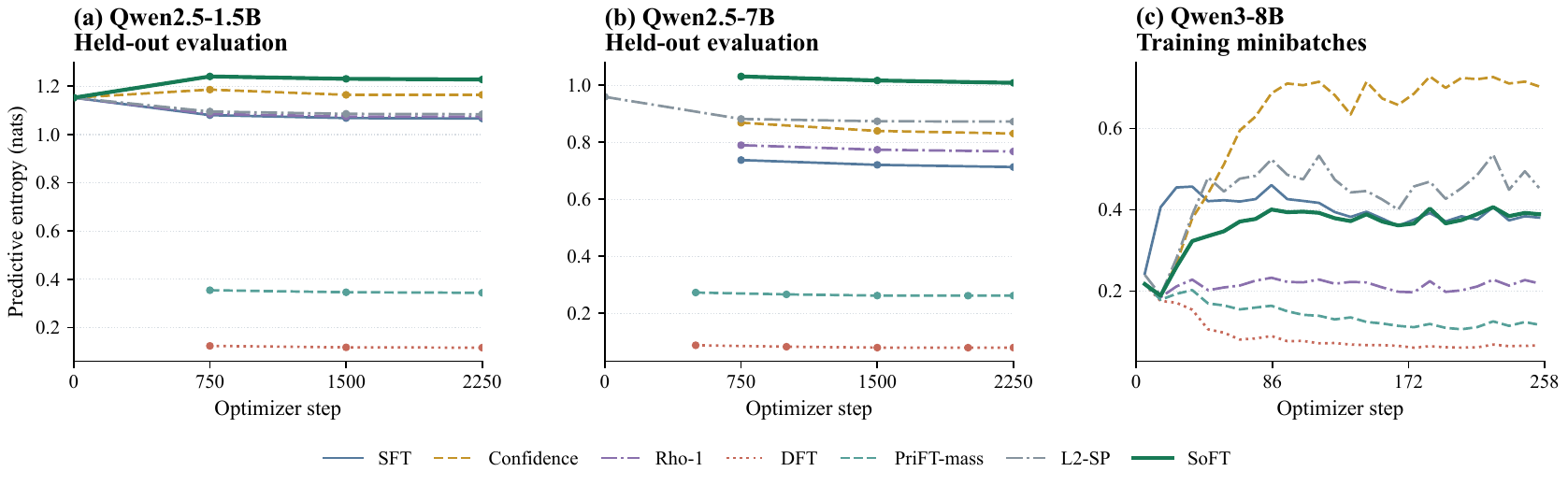}
\caption{Predictive entropy (nats). Left/middle: recorded held-out evaluations. Right: training entropy in 10-step windows. Populations differ across panels; missing points are not imputed.}
\label{fig:training-entropy}
\end{figure}

Figure~\ref{fig:training-entropy} uses held-out entropy for reasoning and training entropy for agentic tasks. Final reasoning entropy is higher for SoFT than SFT: 1.228 versus 1.066 at 1.5B and 1.009 versus 0.713 at 7B. Over the last 10\% of agentic steps, their means are similar (0.395 versus 0.386), while Confidence reaches 0.722 and DFT 0.066. SoFT's entropy therefore depends on the setting. Entropy measures how concentrated predictions are; correctness and Base preservation require separate evaluations. The different populations also limit comparisons across panels.

\subsection{Gradient Norms and Learning-Rate Schedules}

\begin{figure}[!t]
\centering
\includegraphics[width=\linewidth]{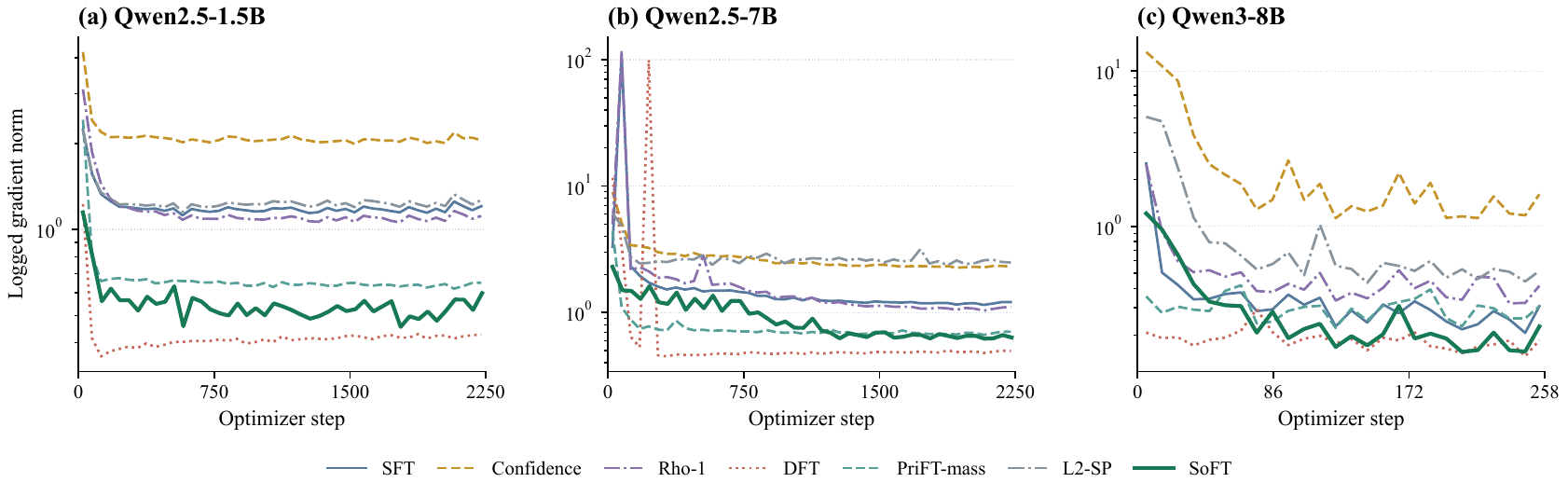}
\caption{Logged gradient norms (log scale), using the windows in Figure~\ref{fig:training-loss}. Magnitudes depend on objective scaling and trainer conventions.}
\label{fig:training-grad-norm}
\end{figure}

SoFT's mean logged gradient norm decreases between the first and last 10\% of steps: 0.770 to 0.555 at 1.5B, 1.659 to 0.644 at 7B, and 1.014 to 0.178 at 8B (Figure~\ref{fig:training-grad-norm}). These values describe the logged update scale, and their interpretation depends on loss scaling, trainer conventions, and fluctuations hidden by window averages.

\begin{figure}[!t]
\centering
\includegraphics[width=\linewidth]{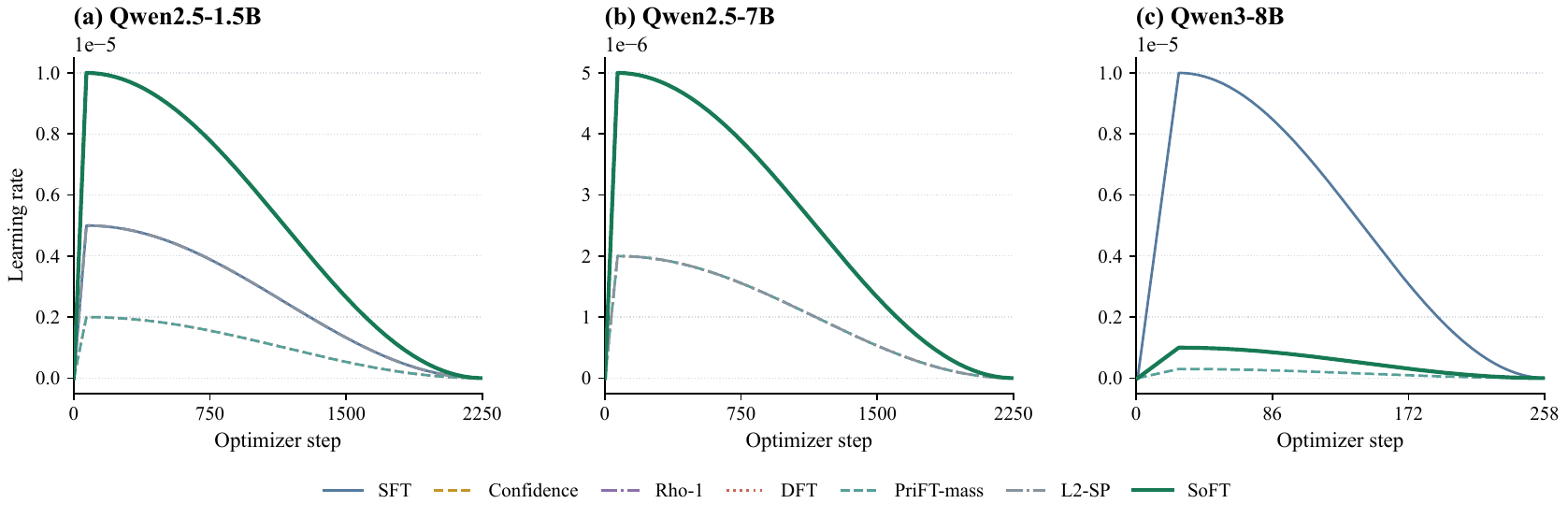}
\caption{Recorded learning-rate schedules without smoothing. Identical schedules overlap.}
\label{fig:training-lr}
\end{figure}

Figure~\ref{fig:training-lr} shows SoFT's peak rates of $10^{-5}$, $5\times10^{-6}$, and $10^{-6}$, followed by decay. Other methods use their own selected rates, and identical schedules overlap. Together, these curves characterize the optimization schedules chosen for the main comparisons.

\section{Realized Demonstration Learning}
\label{app:realized-acquisition}

For each Qwen2.5 scale, we select 32 validation examples per reasoning domain by a fixed hash, using prompts of at most 512 tokens and at least 128 demonstrated tokens. All seven trained methods are scored on the same teacher-forced states and the first 128 demonstrated tokens. We report the gold-token probability gain relative to Base's remaining probability mass, as defined in Section~\ref{sec:explore-acquisition}. Horizontal bars in Figure~\ref{fig:acquisition-performance} show approximate 95\% ranges from 2,000 domain-stratified resamples of the selected trajectories; vertical coordinates are the main-table benchmark GMs. The prefix measurements use validation data, while the task scores come from the separate evaluation suites.

At 7B, three methods with almost the same aggregate acquisition ratio occupy widely separated positions on the vertical axis. Similar gold-token gains can therefore accompany different task outcomes. The scalar summarizes teacher-forced probability gain, while the retention analysis and benchmark scores capture behavior on complete responses.

\begin{table}[!htbp]
\centering
\caption{Realized acquisition by reasoning domain (\%). Values use the same 32 validation examples per domain as Figure~\ref{fig:acquisition-performance}.}
\label{tab:acquisition-by-domain}
\small
\setlength{\tabcolsep}{5pt}
\begin{tabular*}{\linewidth}{@{\extracolsep{\fill}}lrrrrrr@{}}
\toprule
 & \multicolumn{3}{c}{Qwen2.5-1.5B} & \multicolumn{3}{c}{Qwen2.5-7B} \\
\cmidrule(lr){2-4}\cmidrule(l){5-7}
Domain & SFT & DFT & SoFT & SFT & DFT & SoFT \\
\midrule
General & 20.6 & 41.6 & 4.0 & 43.4 & 50.1 & 8.8 \\
Math & 24.9 & 47.2 & 6.0 & 47.2 & 52.9 & 11.3 \\
Science & 18.8 & 37.1 & 7.5 & 36.9 & 44.4 & 11.8 \\
Code & 18.9 & 40.0 & 10.2 & 38.4 & 44.2 & 16.0 \\
\bottomrule
\end{tabular*}
\end{table}

The ordering DFT $>$ SFT $>$ SoFT holds in all four domains at both scales, so the aggregate contrast is not driven by a single domain. Within SoFT, Code has the largest realized gain at both scales and also receives the largest validation-selected budget. The domain budget and Base probability profile jointly shape the target used for these trajectories.

\end{document}